\documentclass[11pt,3p,authoryear]{elsarticle}
\usepackage{lineno,hyperref}
\usepackage{eurosym}
\usepackage{booktabs}
\usepackage{multirow}
\usepackage{graphicx}         
\usepackage{multicol}       
\usepackage{yfonts}
\usepackage{xcolor}
\usepackage{nicefrac}
\usepackage[linesnumbered,ruled,vlined]{algorithm2e}
\usepackage{amsmath,dsfont}
\usepackage{float}
\usepackage{amssymb}
\usepackage[toc,page]{appendix}
\newtheorem{mydef}{Definition}
\newtheorem{as}{Assumption}

\newtheorem{bsp}{Example}
\newtheorem{cor}{Corollary}
\newtheorem{prop}{Proposition}

\SetCommentSty{mycommfont}

\SetKwInput{KwInput}{Input}                
\SetKwInput{KwOutput}{Output}              
\journal{Elsevier}
\begin{document}
\allowdisplaybreaks
\begin{frontmatter}

\title{Information efficient learning of complexly structured preferences:\\
Elicitation procedures and their application to\\ decision making under uncertainty}

\author[rvt]{C.~Jansen\corref{cor1}}
\ead{christoph.jansen@stat.uni-muenchen.de}
\author[rvt]{H. Blocher}
\author[rvt]{T.~Augustin}
\author[rvt]{G.~Schollmeyer}
\cortext[cor1]{Corresponding author}
\address[rvt]{Department of Statistics, LMU Munich, Ludwigsstr. 33, 80539 Munich, Germany}

\begin{abstract}
In this paper we propose efficient methods for elicitation of complexly structured preferences and utilize these in problems of decision making under (severe) uncertainty.  Based on the general framework introduced in Jansen, Schollmeyer \& Augustin (2018, Int. J. Approx. Reason), we now design elicitation procedures and algorithms that enable decision makers to reveal their underlying preference system (i.e.~two relations, one encoding the ordinal, the other the cardinal part of the preferences) while having to answer as few as possible simple ranking questions. Here, two different approaches are followed. The first approach directly utilizes the collected ranking data for obtaining the ordinal part of the preferences, while their cardinal part is constructed implicitly by measuring the decision maker's consideration times. In contrast, the second approach explicitly elicits also the cardinal part of the decision maker's preference system, however, only an approximate version of it. This approximation is obtained by additionally collecting labels of preference strength during the elicitation procedure. For both approaches, we give conditions under which they produce the decision maker's true preference system and investigate how their efficiency can be improved. For the latter purpose, besides data-free approaches, we also discuss ways for statistically guiding the elicitation procedure if data from elicitations of previous decision makers is available. Finally, we demonstrate how the proposed elicitation methods can be utilized in problems of decision under (severe) uncertainty. Precisely, we show that under certain conditions optimal decisions can be found without fully specifying the preference system. 
\end{abstract}


\end{frontmatter}


\section{Introduction}
Making comprehensible and well founded decisions in situations under uncertainty could hardly be more relevant than it is nowadays. 
More and more scientific disciplines are utilizing a decision theoretic framework for (re-)formulating classical tasks of their field or accessing new challenges in an elegant and abstract way that avoids getting distracted with unnecessary details. 
This, in particular, is due to the fact that such framework is both simple and expressive and, accordingly, is perfectly suited for reducing highly complex and inaccessible scientific problems to their absolute core. 
Modern examples for this phenomenon are ranging from applications in personalized medicine (see, e.g.,~\citet{KADI}), climate sciences (see, e.g.,~\citet{held2010,neubersch2014operationalizing}) or environmental management (see, e.g.,~\citet{troffaes}) to applications in statistics (see, e.g.,~\citet{cattaneo, HABLE,cattaneo2013likelihood,patrick}), classification (see, e.g.,~\citet{HULLERMEIER20081897,furnkranz2010preference,aleatoric}) or even quantum mechanics (see, e.g.,~\citet{pmlr-v58-benavoli17a}). 

Having said all this, the following question remains of tremendous relevance: \textit{What makes a decision a good one?} Of course, this question is as old as the theory of decision under uncertainty itself and very elegant and sophisticated answers have been given to it. These answers range from axiomatic characterizations of rational decision making (see, e.g.,~\citet{vnm,foundstat,aumann1,gilboa3,nau}) over more applied approaches (see, e.g.,~\citet{case,benhaim}) to concepts that are still applicable under very general models of uncertainty (see, e.g.,~\citet{kofler,kofler2,Walley.1991,Weichselberger.2001}). A recent discussion challenging the classical interpretation of decision models is given in~\citet{GILBOA2018240}.

From our perspective, there are three main aspects that should be considered when answering this question. We list these now, since we will return to them at several points of the paper.
\begin{itemize}
    \item[(I)] Any meaningful decision should be obtained by some \textit{convincing} decision rule.
    \item[(II)] Applying the chosen formalism for decision making should require only \textit{expertise in the concrete substance matter context} but no strong background in mathematics.
    \item[(III)] Appropriately executing the chosen formalism \textit{should not require more information} than the actual decision, i.e. should be \textit{information efficient}. 

\end{itemize}

Point (I) seems pretty obvious: Why should I base my decision on some rule which I am not convinced of? However, note that even if the rule per se is convincing, it might rely on too idealistic assumptions. For example, even if you agree with the idea of maximizing your expected utility in principle, the criterion loses its bite if you cannot rely on the appropriateness of the involved utility function or probability function or both. So, by convincing we also mean that the theoretical assumptions underlying the decision rule are indeed met. 

Point (II) is also almost self-explaining: Even the most accurate decision rule is useless if it is too technical or complicated to be applied by the decision maker under consideration. Importantly, note that our intention is not to rule out decision criteria that rely on advanced methods. Instead, in our eyes the goal is to present such complex criteria in a way that allows them being applied by practitioners without strong mathematical background. 

Finally, also point (III) seems intuitive: The decision rule should not demand the decision maker to collect information that is not directly relevant for the concrete decision to be made. Otherwise, valuable resources (such as time, money, etc.) would simply be wasted. To give a graphical example: Why should a decision maker specify a precise utility function for all alternatives that are theoretically possible, if for the concrete decision to be made it suffices to have utility intervals for some of the alternatives? Note that a somewhat similar line of argumentation has been followed in~\citet{haddaway}.

We will start our considerations by tackling points (II) and (III) while assuming (I) to be valid. In our eyes, any attempt of doing justice to these demands necessarily starts with the questions how (and how long) the decision maker's preferences over the relevant set of consequences are elicited, since most of the classical criteria for decision making already by definition rely on (too) exactly elicited utility and probability models. In the context of elicitation these demands translate to:
\begin{itemize}
    \item[(II)$^{'}$] The elicitation procedure should be based on simple and understandable questions about the concrete substance matter context.
     \item[(III)$^{'}$] The elicitation procedure should not require more questions than are needed to evaluate the decision rule chosen in (I).
\end{itemize}

The goal of the present paper is now to develop elicitation methods that best possibly approach the demands (II)$^{'}$ and (III)$^{'}$ and then utilizing these procedures in decision making problems for different choices of the decision rule (I). Note that there is a vast amount of literature on elicitation of preferences and utility (see, e.g., \citet{eugene},~\citet{abdel}~or~\citet{fischhoff} to only name a few). The references that seem most relevant in our context are~\cite{danielson},~\cite{danisipta}~and~\cite{pmlr-v62-troffaes17b} as these, similar to us, also allow for partial preferences and investigate connections to decision making under imprecise probabilities. For a discussion of relations and differences of these and our approach see~\cite{jsa2018}. For a recent discussion of the topic from a rather philosophical point of view see, e.g.,~\citet{baccelli2016choice}.

Our paper is organized as follows: We start by recalling the required mathematical concepts in Section~\ref{prel}. In Section~\ref{elisec}, we propose two different procedures for eliciting a decision maker's preference system by only asking simple ranking questions. The first procedure (Section~\ref{pro1}) utilizes the decision maker's consideration times, whereas the second procedure (Section~\ref{pro2}) instead asks for labels of preference strength. Section~\ref{discussion} provides a discussion of the assumptions underlying Procedure 1. For both procedures, we demonstrate possibilities for improving their efficiency. Besides data-free approaches, Section~\ref{data} also discusses ways  for  effectively  guiding  the  elicitation  procedure  if  data  from  previous  elicitation  rounds  is available. Section~\ref{decmak} builds the bridge to decision making under uncertainty: After Section~\ref{bm} recalled the basic decision theoretic framework, Section~\ref{rule} presents two decision rules that, in our eyes, meet demand (I). Section~\ref{dmel} demonstrates how to use the elicitation procedures from Section~\ref{elisec} for more efficient decision making. Section~\ref{aex} discusses two examples of a selection of the concepts discussed. Section~\ref{outlook} concludes the paper by elaborating on some promising perspectives for future research.
\section{Executive  summary  for  a  non-mathematical  audience}

In order to improve exchange with potential practical users of our approach, we first give a non-technical account to the main ideas of our work, to be developed rigorously in the following sections. Our aim is to provide a formal basis for building up a practical decision support system in which users are guided to optimal decisions properly reflecting their individual preferences. To be powerfully usable in the intended applications, the system has to be flexible and information efficient. Flexibility shall mean that there the user is not forced to formulate utility values, or indirectly preferences, of unrealistically high precision. This is mathematically achieved by our {\em preference systems} that allow for incomplete rankings and indecisiveness between the consequences to rank (see later on relation $R_1$) and a user-friendly, comparative expression of preference strengths (leading to relation $R_2$), either by eliciting them indirectly utilizing data from the ranking process itself (Section \ref{pro1}) or directly from a natural ordinal scale (Section \ref{pro2}). Information efficiency shall guarantee that the elicitation process parsimoniously restricts itself to as few ranking questions as are needed to arrive at an optimal decision.

In what follows, we want to obtain a preference system on a finite set of consequences. On the one hand, we need to collect the information on the decision maker's preferences between the consequences themselves, allowing that two consequences are incomparable or can be ranked by preference in either direction. On the other hand, to fully specify the preference system, we want to obtain the information whether preferences on the set of all comparisons of consequences exist. More precisely, this means that it may also contain information on, e.g., whether exchanging consequence $b$ by consequence $a$ is at least as desirable as exchanging consequence $d$ by consequence $c$. Thus, in our elicitation procedures, we want to derive not only whether there is a preference between two consequences and in which direction it goes, but also the strength of the corresponding preference. Therefore, the complete specification of the preference system can be very time consuming even for a small number of consequences. The approaches presented here are based on the idea that preference strength is not asked directly, but is encoded in the question of which of the two 
consequences is preferred. 

In the \textit{time elicitation} method (Section \ref{pro1}), we obtain the preference strength by using the consideration time of the decision maker for ranking the 
two consequences. This gives us a ranking system on the set of all comparisons. This procedure assumes that the preference strength and the consideration 
time are related in the sense that the preference is high when the decision maker decides quickly. Thus, the ranking of the preference system is implicitly given here. In contrast, in the second procedure, called \textit{label elicitation} (Section \ref{pro2}), we explicitly elicit labels of preference strength. Again, we ask the 
decision maker which consequence out of two is preferred, and the decision maker then additionally assigns a label from a predetermined set of labels. 
The main assumption now is that the label refers to the preference strength. Algorithm~\ref{time elicitation} and 
Algorithm~\ref{hierlabel} are efficient modifications of the \textit{time elicitation} and \textit{label elicitation} methods, respectively. 
Note that these algorithms have further technical assumptions and/or assume that the preference system must be transitive. 
Furthermore, if we already have a sample of the preference systems of previous users and the assumption that the decision maker has a similar preference system is true, then we can order the asked question, i.e., which consequence from a pair is preferred, in a \textit{statistically guided} way (Section \ref{data}).

In the last part of the paper (Section \ref{decmak}), we investigate what can be learned from our elicitation procedures for problems of decision making \textit{under uncertainty}. Turning to the additional incorporation of uncertainty, we no longer consider just a single set of consequences, but actually $acts$ that map from a set of uncertain states of nature to a subset of all consequences. 
This allows us to add prior knowledge about the uncertainty on the states using either precise or imprecise probabilistic models. It turns out that, under certain 
conditions, determining an optimal act in a decision problem under uncertainty can then be done in a very information efficient way by recursively applying either the (modified) 
\textit{time} or \textit{label elicitation} procedure.

\section{Preliminaries}\label{prel}
Throughout the paper, we assume $A=\{a_1 , \dots , a_n\}$ to be a finite set of consequences. The elements of $A$ are interpreted as all potential consequences that the choices of a decision maker could possibly lead to. For clarity it should be mentioned that $A$ should not be misinterpreted as the set of available acts in a decision making problem under uncertainty. As far as only $A$ is concerned, there is no (probabilistic) uncertainty involved. The connection to decision making under uncertainty becomes clear in Section~\ref{decmak}: Here we consider the problem of choosing between acts $X:S \to A$ taking values in such a set $A$. The uncertainty is then about which of the states of nature collected in $S$ corresponds to the true description of reality. 

As part of our main focus lies on eliciting the decision maker's preferences on the consequence set $A$, we start by recalling some basic notions of binary relations. For a binary relation $R \subseteq M \times M$ on a set $M$ (which will often be the set $A$ itself in what follows), we denote by $P_R \subseteq M \times M$ the \textit{strict part} of $R$ defined by 
\begin{equation}
(m_1 , m_2) \in P_{R} \Leftrightarrow (m_1 , m_2) \in R \wedge (m_2 , m_1) \notin R
\end{equation}
by $I_R \subseteq M \times M$ the \textit{indifference part} of $R$ defined by 
\begin{equation}
(m_1 , m_2) \in I_{R} \Leftrightarrow (m_1 , m_2) \in R \wedge (m_2 , m_1) \in R
\end{equation}
by $C_R \subseteq M \times M$ the \textit{incomparable part} of $R$ defined by 
\begin{equation}
(m_1 , m_2) \in C_{R} \Leftrightarrow (m_1 , m_2) \notin R \wedge (m_2 , m_1) \notin R
\end{equation}
and by $R^{-1} \subseteq M \times M$ the \textit{inverse relation} of $R$ defined by
\begin{equation}
(m_1 , m_2) \in R^{-1}\Leftrightarrow (m_2 , m_1) \in R.
\end{equation}
Observe that $R$ is the disjoint union of $I_R$ and $P_R$ and that $M \times M$ is the disjoint union of $P_R$, $P_R^{-1}$, $I_R$ and $C_R$. Moreover, the \textit{transitive hull} of the relation $R$ is given by
\begin{equation}
H(R)=
\Bigl\{(a,b)\Bigl|\small{\begin{tabular}{l}$\exists k \in \mathbb{N} ~ \exists m_1, \dots , m_k \in M: m_1=a  ~\wedge$ \\ $m_k=b~\wedge ~ \forall i<k: (m_{i}, m_{i+1}) \in R$ \\\end{tabular}}\Bigr\}   
\end{equation}
i.e.~by the set of pairs $(a, b) \in M \times M$ that can be connected by a path of finite length in the graph associated with $R$. Note that computation of $H(R)$ is feasible (cf., \citet{Fischer1971}).

As our main tool for modelling a decision maker's preferences on $A$, we will make use of \textit{preference systems} as introduced in~\cite{jsa2018}. These allow to model preferences that are partially ordinal and partially cardinal and provide a very general and expressive formal framework.
\begin{mydef} Let $A$ be a non-empty set and let $R_1 \subseteq A \times A$ denote a binary relation on $A$. Moreover, let $R_2 \subseteq R_1 \times R_1$ denote a binary relation on $R_1$. Then the triplet $\mathcal{A}=[A, R_1 , R_2]$ is called a \textbf{preference system} on $A$. Moreover, a preference system $\mathcal{A}^{'}=[A, R^{'}_1 , R^{'}_2]$ is called \textbf{sub-system} of $\mathcal{A}$ if $R^{'}_1\subseteq R_1$ and $R^{'}_2\subseteq R_2$.
\label{ps}
\end{mydef}
Some words on the interpretation of preference systems: The \textit{ordinal part} of the agent's preferences is formalized by some binary relation $R_1 \subseteq A \times A$. Importantly, note that $R_1$ does not need to be complete, that is there might exist elements $a,b \in A$ for which both $(a,b) \notin R_1$ and $(b,a) \notin R_1$ holds. If we have that $(a,b) \in R_1$, then we interpret this as $a$ being \textit{at least as desirable as} $b$ for the decision maker under investigation. The induced relations $I_{R_1}$ and $P_{R_1}$ are interpreted as indifference and strict preference, respectively.

The \textit{cardinal part} of the agent's preferences is formalized by some $R_2 \subseteq R_1 \times R_1$, i.e. a binary relation between that pairs of consequences that are in relation with respect to the relation $R_1$. If a pair of pairs satisfies $((a,b),(c,d)) \in R_2$, then we interpret this as \textit{exchanging $b$ by $a$ being at least as desirable as exchanging $d$ by $c$}. The relations $I_{R_2}$ and $P_{R_2}$ are interpreted as indifference and strict preference between such exchanges, respectively. Note that also $R_2$ does not need to be complete such that there might exist exchanges of consequences which are incomparable for the decision maker.

As the definition of a preference system does not restrict the relations $R_1$ and $R_2$ at all (note that not even transitivity is required), the following definition introduces a concept for distinguishing rational and irrational preference systems. It is taken from~\citet[Definition 2]{jsa2018}.
\begin{mydef} Let $\mathcal{A}=[A, R_1 , R_2]$ be a preference system. Then $\mathcal{A}$ is said to be \textbf{consistent} if there exists a function $u:A \to [0,1]$ such that for all $a,b,c,d \in A$ the following properties hold:
\begin{itemize}
\item[i)] If $ (a , b) \in R_1$, then $u(a) \geq u(b)$ with equality iff $(a,b)\in I_{R_1}$.
\item[ii)] If $((a , b),(c,d)) \in R_2$, then $u(a) -u(b) \geq u(c)-u(d)$ with equality iff $((a,b),(c,d))\in I_{R_2}$.
\end{itemize}
Every such function $u$ is then said to \textbf{(weakly) represent} the preference system $\mathcal{A}$.
The set of all (weak) representations $u$ of $\mathcal{A}$ is denoted by $\mathcal{U}_{\mathcal{A}}$. 
\label{consistent}
\end{mydef}
Since both the generalized expectation interval from Definition~\ref{gei} and the decision rule from Definition~\ref{cf} i) are not directly based on $\mathcal{U}_{\mathcal{A}}$ itself but on sets $\mathcal{N}_{\mathcal{A}}$ or $\mathcal{N}^{\delta}_{\mathcal{A}}$ instead, we briefly describe their construction:\footnote{The construction of the sets $\mathcal{N}_{\mathcal{A}}$ and $\mathcal{N}^{\delta}_{\mathcal{A}}$ used here is a slight variation (but essentially the same) as the one given in \citet{jsa2018}, Definitions~2~and~3, respectively.} Define an equivalence relation $\sim$ on $\mathcal{U}_{\mathcal{A}}$ by setting 
\begin{equation}
(u_1, u_2) \in \sim ~~~\Leftrightarrow ~~~ \exists: m \in \mathbb{R}^+, t \in \mathbb{R} \text{ such that } u_1 = m \cdot u_2+t.
\end{equation}
That is two utility function are equivalent whenever they are positive linear transformations of each other. The corresponding quotient space is then defined by 
\begin{equation}
   \mathcal{U}_{\nicefrac{{\mathcal{A}}}{\sim}}:= \{[u]_{\sim}: u \in \mathcal{U}_{\mathcal{A}}\}
\end{equation}
where $[u]_{\sim}$ denotes the set of all $v \in \mathcal{U}_{\mathcal{A}} $ that are equivalent to $u$.  For defining the sets $\mathcal{N}_{\mathcal{A}}$ and $\mathcal{N}^{\delta}_{\mathcal{A}}$, we need to assume that there exist elements $a_*, a^* \in A$ such that $(a^*,a) \in R_1$ and $(a,a_*) \in R_1$ for all $a \in A$.\footnote{Note that this assumption is only needed for the choice function from Definition~\ref{cf} i) to be well-defined (compare also the discussion directly following Definition~\ref{cf}). All parts of the paper not involving this choice function (in particular all of Section \ref{elisec} and the parts of Section \ref{decmak} concerning Definition~\ref{cf} ii)) avoid this assumption. } Observe that we then collect exactly one representative $r([u]_{\sim})$ from each class $[u]_{\sim}$ by demanding $r([u]_{\sim})(a^*)=1$ and $r([u]_{\sim})(a_*)=0$ . Denote by 
\begin{equation}
    \mathcal{N}_{\mathcal{A}}:= \Bigr\{r([u]_{\sim}): [u]_{\sim} \in  \mathcal{U}_{\nicefrac{{\mathcal{A}}}{\sim}}\Bigl\}
\end{equation}
the set of all such representatives. The set $ \mathcal{N}_{\mathcal{A}}$ can be thought of as all compatible utility functions that measure utility on a $[0,1]$-scale.

 Finally, for a number $\delta \in [0,1)$, we denote by $\mathcal{N}^{\delta}_{\mathcal{A}}$ the set of all $u \in \mathcal{N}_{\mathcal{A}}$ satisfying 
 $$ u(a)-u(b) \geq \delta ~~~\wedge ~~~ u(c)-u(d) -u(e) + u(f) \geq \delta $$ 
 for all $(a,b) \in P_{R_1}$ and for all $((c,d),(e,f)) \in P_{R_2} $. The parameter $\delta$ is called \textit{granularity} and the preference system $\mathcal{A}$ is called $\mathbf{\delta}$-consistent if $\mathcal{N}_{\mathcal{A}}^{\delta} \neq  \emptyset$. Very roughly, $\delta$ can be interpreted as the minimal utility difference the decision maker wants to account for in a practical problem. For more details on the interpretation of $\delta$ see the discussion after Definition 3 in \citet{jsa2018}.
\section{Elicitation of preference systems} \label{elisec}
From now on relations $R_1^*$ and $R_2^*$ marked with an asterisk always correspond to the decision maker's true relations, whereas $R_1$ and $R_2$ denote the relations obtained by some elicitation procedure. Note that throughout the rest of the paper the relations $R_1^*$ and $R_2^*$ are assumed to be reflexive. We can now formulate our central question: How can we elicit a decision maker's true preference system $\mathcal{A}^*=[A,R_1^*,R_2^*]$ ? Clearly, an immediate strategy would be directly asking for all preferences. But while this strategy seems to be feasible for the ordinal part $R^*_1$ of the preference system at least for a moderate number $n$ of consequences, it becomes rapidly infeasible for its cardinal part $R^*_2$. For instance, even for a relatively small number of consequences such as $n=10$, the decision maker would have to answer $45$ questions to fully specify $R^*_1$ and $4950$ questions to fully specify $R^*_2$.\footnote{Observe that, without further refinements, fully specifying $R^*_1$ requires asking up to $\tfrac{n(n-1)}{2}$ questions while fully specifying $R^*_2$ requires asking up to $\tfrac{n^2(n^2-1)}{2}$ questions.} The number of questions to ask increases of order $\Theta(n^4)$ in the number $n$ of different consequences. Thus, simply asking for all preferences seems not to be reasonable.

In the following two subsections, we propose two procedures that avoid directly eliciting the cardinal part $R^*_2$ of the preference system and instead implicitly construct it by utilizing subject-specific information about the elicitation process of $R_1^*$ itself. Specifically, the first procedure relies on \textit{consideration times} of the decision maker, while the second procedure additionally collects \textit{labels of preferences strength} for every pair $(a_i,a_j)$. We then show that, under suitable conditions, efficiency of both procedures can be further improved.
\subsection{Procedure 1: Time elicitation} \label{pro1}

Our first procedure for eliciting the preference system $\mathcal{A}^*=[A,R_1^*,R_2^*]$ of some decision maker relies on \textit{data about consideration times} that is obtained during the elicitation process of $R^*_1$. The intuition behind it is that the strength of the preference between two consequences \textit{decreases} in the time that ranking the two consequences takes. In particular, for each pair of alternatives $(a_i,a_j) \in P_{R_1^*}$, besides the ranking, we also measure the consideration time $t_{ij}$ the decision maker needs for ranking the two consequences. This view is a powerful transfer of the paradata approach
from survey methodology (e.g.~\citet{Kreuter:2013}), 
based on the insight that individual data about the surveying process itself can contain valuable information about respondents and their dispositions. Consequently, the consideration times are then utilized for constructing a cardinal relation $R_2$ being a candidate for the true $R_2^*$. This constructs a preference system while asking only questions on the decision maker's ordinal preferences $R^*_1$. \\[.15cm]
\indent\textbf{Time elicitation}\footnote{Some very early-staged ideas in the context of time elicitation were already discussed in~\citet[p. 31]{jansen2018some}.} can be described as follows: As before, we have a finite set $A=\{a_1, \dots , a_n\}$. 
We start with three initial relations $R_1 = \{(a,a):a \in A\}$ and $R_2 = \emptyset $ and $C= \emptyset$. We now successively ask the decision maker about the preferences between certain (not necessarily all) pairs $\{a_i , a_j\}$ from the set $$ A_{\{2\}}:=\{\{a,b\}:a \neq b \in A\}.$$ 
There are four possibilities:
\begin{itemize}
\item[i)] The decision maker judges $a_i$ and $a_j$ to be incomparable. In this case $R_1$ and $R_2$ remain unchanged and we add $(a_j , a_i)$ and $(a_i , a_j)$ to $C$. The times $t_{ij}$ and $t_{ji}$ are set to zero.
\item[ii)]  The decision maker ranks $a_i$ strictly more preferable than $a_j$. In this case we add $(a_i , a_j)$ to $R_1$ and measure the time $t_{ij} >0$. The time $t_{ji}$ is set to zero.
\item[iii)]  The decision maker ranks $a_j$ strictly more preferable than $a_i$. In this case we add $(a_j , a_i)$ to $R_1$ and measure the time $t_{ji}>0$. The time $t_{ij}$ is set to zero.
\item[iv)]  The decision maker is indifferent between $a_j$ and $a_i$. We then add the pairs $(a_j , a_i)$ and $(a_i , a_j)$ to $R_1$. The times $t_{ij}$ and $t_{ji}$ are set to some $c_{\infty} \in \mathbb{R}$ with $c_{\infty}>  \max\{t_{pq}: (a_p,a_q) \in P_{R_1^*}\}$.
\end{itemize} 
This leaves us with a relation $R_1$ approximating $R^*_1$. Next, we first set $t_{ii}:=c_{\infty}$ for all $i=1 , \dots, n$ and then utilize the consideration times for constructing an approximation $R_2$ of $R_2^*$. For that, we successively pick pairs of pairs $(a_i , a_j)$, $(a_k , a_l) \in R_1$ and define $R_2$ by
\begin{equation}
    ((a_i,a_j),(a_k,a_l)) \in R_2~~~:\Leftrightarrow~~~t_{kl}-t_{ij}\geq 0 ~\wedge~t_{ij}>0
\end{equation}
that is if the decision between $a_k$ and $a_l$ took at least as long as the decision between $a_i$ and $a_j$. Finally, this procedure produces a preference system $\mathcal{A}=[A,R_1,R_2]$ on $A$.\\[.15cm] 
\indent Importantly, note that the preference system produced by time elicitation is only meaningful if one accepts the following assumption on the relation between the decision maker's consideration times and the true preferences system $\mathcal{A}^*$:\footnote{We will discuss the appropriateness and the limitations of this and the other assumptions underlying time elicitation in Section~\ref{discussion} and show how to deal with situations in which these assumptions are not (even approximately) satisfied in Section~\ref{pro2}.} 
\begin{as} \label{time assumption}
For $(a_i , a_j)$, $(a_k , a_l) \in R^*_1$ the following holds true: 
\begin{itemize}
    \item[i)] $t_{kl}>t_{ij}>0 $ if and only if $((a_i , a_j),(a_k , a_l)) \in P_{R^*_2}$.
     \item[ii)] $t_{kl}=t_{ij}>0 $ if and only if $((a_i , a_j),(a_k , a_l)) \in I_{R^*_2}$.
     \item[iii)] The maximal consideration time $c_{\infty}$ is exactly attained for indifferent consequences, i.e.\\ $t_{ij}=t_{ji}=c_{\infty}$ if and only if $(a_i,a_j) \in I_{R^*_1}$.
\end{itemize}
\end{as}
The following proposition states that Assumption~\ref{time assumption} is enough to guarantee that time elicitation indeed produces the true preference system if all pairs of consequences are presented.
\begin{prop}\label{prop1}
Let Assumption~\ref{time assumption} hold true. Then time elicitation produces the decision maker's true preference system $\mathcal{A}^*=[A,R_1^*,R_2^*]$, if every element from $A_{\{2\}}$ is presented.
\end{prop}
\textit{Proof.} Let $\mathcal{A}=[A,R_1,R_2]$ be the preference system produced by time elicitation after having shown all pairs from $A_{\{2\}}$. We must show (1)~$R_1=R_1^*$ and (2) $R_2=R_2^*$. Equation~(1) directly follows by definition of time elicitation utilizing that $R_1^*$ is reflexive and the fact that every pair has been presented. To see equation (2), choose $((a_i , a_j),(a_k , a_l)) \in R_2$. By definition, this implies $t_{kl}-t_{ij}\geq 0 ~\wedge~t_{ij}>0$. There are two cases:\\[.1cm]
\textit{Case 1:} $t_{kl}=t_{ij}>0$. This implies $((a_i , a_j),(a_k , a_l)) \in I_{R^*_2}$ according to Assumption~\ref{time assumption} ii).\\[.1cm]
\textit{Case 2:} $t_{kl}>t_{ij}>0$. This implies $((a_i , a_j),(a_k , a_l)) \in P_{R^*_2}$ according to Assumption~\ref{time assumption} i).\\[.1cm]
Thus, we have $((a_i , a_j),(a_k , a_l)) \in R^*_2$. Let conversely $((a_i , a_j),(a_k , a_l)) \in R^*_2$. First, by Assumption~\ref{time assumption} i) and ii), this implies $t_{kl}-t_{ij}\geq 0 ~\wedge~t_{ij}>0$. Second, since by definition of $R_2^*$ we know that $(a_i , a_j),(a_k , a_l) \in R^*_1$, this also implies $(a_i , a_j),(a_k , a_l) \in R_1$ by (1). Together this yields $((a_i , a_j),(a_k , a_l)) \in R_2$, completing the proof.
\hfill $\square$
\begin{cor} \label{cor1}
Let Assumption~\ref{time assumption} hold true. Then time elicitation produces a sub-system of $\mathcal{A}^*$ for any presented subset $B \subseteq A_{\{2\}}$. This sub-system is consistent whenever $\mathcal{A}^*$ is.\hfill $\square$
\end{cor}
Importantly, observe that the elicitation procedure not necessarily produces preferences systems that are consistent in the sense of Definition~\ref{consistent}. For an immediate counterexample just consider a decision maker with intransitive ordinal preferences, e.g., $(a,b) \in P_{R_1^*}$, $(b,c) \in P_{R_1^*}$ and $(c,a) \in P_{R_1^*}$. Note that checking consistency of the produced preference system can be done by solving one single linear optimization problem, see~\cite[Proposition 1]{jsa2018}. 

The following proposition states that transitivity of the ordinal part $R_1^*$ of the preference system is sufficient for time elicitation to produce consistent preference systems, if two further assumptions on the connection between consideration times and preferences strength are satisfied: 
\begin{as} \label{additive}
For $(a_i , a_j)$, $(a_j , a_k) \in P_{R^*_1}$ we have $\tfrac{1}{t_{ij}}+\tfrac{1}{t_{jk}}=\tfrac{1}{t_{ik}}$, whenever $(a_i , a_k) \in P_{R^*_1}$.
\end{as}
\begin{as} \label{indi}
For $(a_i , a_j)\in I_{R^*_1}$ we have
\begin{itemize}
\item[i)] $t_{ki}=t_{kj}$ whenever $(a_k,a_i), (a_k,a_j) \in P_{R_1^*}$ and
\item[ii)] $t_{ik}=t_{jk}$ whenever $(a_i,a_k), (a_j,a_k) \in P_{R_1^*}$.
\end{itemize}
\end{as}
Given the additional assumptions just stated, we indeed can show that transitivity of the decision maker's ordinal preferences guarantees the underlying preference system to be consistent.
\begin{prop} \label{consistent1}
Under the Assumptions~\ref{time assumption},~\ref{additive} and~\ref{indi} the decision maker's true preference system $\mathcal{A}^*=[A,R^*_1,R^*_2]$ is consistent if and only if $R_1^*$ is transitive.
\end{prop}
\textit{Proof.} First, observe that if $\mathcal{A}^*$ is consistent, then $R_1^*$ obviously needs to be transitive. For the converse direction, assume $R_1^*$ to be transitive. We need to show that there exists a function $u:A \to [0,1]$ that weakly represents $\mathcal{A}^*$ in the sense of Definition~\ref{consistent}. Without loss of generality, we assume that $a_1 \in A$ is chosen such that $(a_1,a_i) \in P_{R_1^*}$ for all $i=2, \dots ,n$.\footnote{Observe that if there does not exist an element which is strictly preferred to all other consequences, we can simply add such an element before beginning the elicitation procedure.\label{foot6}} Now, set $u(a_1):=1$. Moreover, for all $i=2, \dots ,n$, we set $u(a_i):=1-\tfrac{1}{t_{1i}}$. We start by showing that this function $u$ represents the relation $R_1^*$. To see that, let $(a_i,a_j) \in R_1^*$ be arbitrary. We distinguish four cases:\\[.1cm]
\textit{Case 1:} $(a_i,a_j) \in P_{R_1^*}$ and $a_i=a_1$. Then $u(a_i)-u(a_j)=1-(1-\tfrac{1}{t_{1j}})=\tfrac{1}{t_{1j}}>0$.\\[.1cm]
\textit{Case 2:} $(a_i,a_j) \in P_{R_1^*}$ and $a_i \neq a_1$. Then $u(a_i)-u(a_j)=\tfrac{1}{t_{1j}}-\tfrac{1}{t_{1i}}$. Since $(a_1,a_i),(a_i,a_j) \in P_{R_1^*}$, it follows by transitivity of $R_1^*$ and Assumption~\ref{additive}, that it holds  $\tfrac{1}{t_{1i}}+\tfrac{1}{t_{ij}}=\tfrac{1}{t_{1j}}$. Together, this yields $u(a_i)-u(a_j)= \tfrac{1}{t_{ij}}>0$.\\[.1cm]
\textit{Case 3:} $(a_i,a_j) \in I_{R_1^*}$ and $a_i=a_1$. This implies that $a_j=a_1$ and, therefore, $u(a_i)-u(a_j)=0.$\\[.1cm]
\textit{Case 4:} $(a_i,a_j) \in I_{R_1^*}$ and $a_i \neq a_1$. Then, by Assumption~\ref{indi}, we can conclude that $t_{1i}=t_{1j}$ and thus $u(a_i)-u(a_j)=\tfrac{1}{t_{1j}}-\tfrac{1}{t_{1i}}=0$.\\[.1cm]
This shows that $u$ represents $R_1^*$. To see that $u$ also represents $R_2^*$ in the sense of Definition~\ref{consistent} ii), choose $((a_i,a_j),(a_k,a_l)) \in R_2^*$ arbitrary. Again, we have to distinguish several cases:\\[.1cm]
\textit{Case 1:} $((a_i,a_j),(a_k,a_l)) \in I_{R_2^*}$. According to Assumption~\ref{time assumption} ii) this implies $t_{ij}=t_{kl}$ and, therefore, we can conclude that $u(a_i)-u(a_j)=\tfrac{1}{t_{ij}}=\tfrac{1}{t_{kl}}=u(a_k)-u(a_l)$.\\[.1cm]
\textit{Case 2:} $((a_i,a_j),(a_k,a_l)) \in P_{R_2^*}$.\\[.1cm]
\textit{Sub-case 2.1:} $(a_i,a_j),(a_k,a_l) \in P_{R_1^*}$. According to Assumption~\ref{time assumption} i) this implies $t_{kl} > t_{ij}$ and, therefore, we can conclude that $u(a_i)-u(a_j)=\tfrac{1}{t_{ij}}>\tfrac{1}{t_{kl}}=u(a_k)-u(a_l)$.\\[.1cm]
\textit{Sub-case 2.2:} $(a_i,a_j) \in P_{R_1^*}$ and $(a_k,a_l) \in I_{R_1^*}$. Then according to Assumptions~\ref{time assumption} i) and iii) we have that $u(a_i)-u(a_j)=\tfrac{1}{t_{ij}}>\tfrac{1}{c_{\infty}}=u(a_k)-u(a_l)$.\\[.1cm]
Finally, observe that the sub-cases with $(a_i,a_j),(a_k,a_l) \in I_{R_1^*}$ or $(a_i,a_j) \in I_{R_1^*}$ and $(a_k,a_l) \in P_{R_1^*}$ are not possible, since they directly conflict with Assumption~\ref{time assumption} and the assumption that $((a_i,a_j),(a_k,a_l)) \in P_{R_2^*}$. This completes the proof.
\hfill $\square$\\[.2cm]
As an immediate consequence of Corollary~\ref{cor1}~and Proposition~\ref{consistent1}, we receive the following statement with conditions for time elicitation to produce consistent sub-systems.
\begin{cor}
Let Assumptions~\ref{time assumption},~\ref{additive} and~\ref{indi} hold true. Then time elicitation produces a consistent sub-system of $\mathcal{A}^*$ for any subset $B \subseteq A_{\{2\}}$ if and only if $R_1^*$ is transitive. \hfill $\square$
\end{cor}
We now show that efficiency of time elicitation can be improved if we assume decision makers with transitive $R_1^*$ whose consideration times satisfy Assumptions~\ref{time assumption},~\ref{additive} and~\ref{indi}. Suppose time elicitation has produced the relations $R_1^{k}$ and $C^{k}$ after $k$ pairs have been presented. Transitivity of $R_1^*$ then allows to \textit{deduce} all preferences for pairs in $H_k\setminus R_1^k$, where $H_{k}=H(R_1^{k})$. The pair to present in step $k+1$ thus may be selected\footnote{One could select the next pair to present simply by sampling randomly. More sophisticated and efficient ways of selecting the next pair are described in Section \ref{data}.}  from the (usually remarkably) smaller set 
\begin{equation*}
A_{\{2\}}\setminus \bigl\{\{a,b\}: (a,b) \in H_{k} \vee (b,a)\in H_{k} \vee (a,b) \in C^k\bigr\}.
\end{equation*}
For also guaranteeing a correct and complete construction of $R_2$,  we need a method for also deducing the decision maker's consideration time $t_{ij}$ for all pairs $(a_i , a_j) \in H_k\setminus R_1^k$, i.e.~all pairs that are not directly presented but deduced from transitivity. Under Assumptions~\ref{time assumption},~\ref{additive} and~\ref{indi}, this is straightforward: For a pair $(a_i , a_j) \in H_k\setminus R_1^k$ such that  $(a_j , a_i)\in H_k$, we can conclude $(a_i , a_j)\in I_{R^*_1}$ and set $t_{ij}$ to $c_{\infty}$ according to Assumption~\ref{time assumption} iii). In contrast, if we have $(a_j , a_i)\notin H_k$, then we can conclude that $(a_i , a_j)\in P_{R^*_1}$, i.e. that we have strict preference in the true preference system. 

First, suppose that $a_i$ and $a_j$ are connected by a path of length $2$ in $R_1^k$, i.e. that there exists some element $a_l \in A$ such that $(a_i , a_l) \in  R_1^k$ and $(a_l , a_j) \in R_1^k$. Note that both $(a_i , a_l) \in  I_{R_1^*}$ and $(a_l , a_j) \in I_{R_1^*}$ is impossible, since we have $(a_i , a_j)\in P_{R^*_1}$. If $(a_i , a_l) \in  P_{R_1^*}$ and $(a_l , a_j) \in I_{R_1^*}$, we know by Assumption~\ref{indi} i) that $t_{ij}$ equals $t_{il}$, which has been measured during the procedure. If $(a_i , a_l) \in  I_{R_1^*}$ and $(a_l , a_j) \in P_{R_1^*}$, we know by Assumption~\ref{indi} ii) that $t_{ij}$ equals $t_{lj}$, which has been measured during the procedure.  This leaves the case  $(a_i , a_l) \in  P_{R_1^*}$ and $(a_l , a_j) \in P_{R_1^*}$. Here, we can utilize Assumption~\ref{additive} to compute the missing consideration time from consideration times collected during the procedure. Simple arithmetic yields
\begin{equation} \label{fill up}
  t_{ij}=   \frac{t_{il}\cdot t_{lj}}{t_{il}+t_{lj}}.
\end{equation}
Importantly, note that the computed value for the missing consideration times does not depend on the choice of the path in $R_1^k$ by Assumption~\ref{additive}. Hence, the procedure is well-defined (for paths of length $2$). Next, observe that the procedure just described can be extended to paths of arbitrary (finite) length $p$ in $R_1^k$: We simply divide the corresponding path into paths of length at most $2$ and use the procedure just described to compute the times for these paths. Afterwards, we receive a path of length less than $p$
 for which we know the consideration times. Now we can divide this path into paths of length at most $2$ and compute their times. Repeating this will end up in a path of length $1$ for which we can compute the consideration time. Clearly, the computed time values are independent of the choice of the original path in $R_1^k$ due to Assumptions~\ref{additive} and~\ref{indi} (as seen above). Additionally, observe that actually in no elicitation step we will have to consider paths of length more than $3$: As $R_1^{k-1}$ is already transitive by construction, for every pair $(a_i , a_j) \in H_k\setminus R_1^k$ there will always exist a path of length at most $3$ in $R_1^k$ connecting the two. A more detailed description of this efficient version of time elicitation for the case that all pairs not implied by transitivity are presented, is given in Algorithm~\ref{time elicitation}. The next proposition states that the efficient version of time elicitation indeed works.
\begin{prop}
Under Assumptions~\ref{time assumption},~\ref{additive} and~\ref{indi}, Algorithm~\ref{time elicitation} terminates in $\mathcal{A}^*$ if and only if $R^*_1$ is transitive. According to Proposition~\ref{consistent1} we know $\mathcal{A}^*$ is consistent in this case.
\end{prop}
\textit{Proof.} Assume that Algorithm~\ref{time elicitation}~has terminated in the preference system $\mathcal{A}=[A,R_1,R_2]$. By construction, the relation $R_1$ is transitive. Thus, the relation $R_1^*$ needs to be transitive whenever Algorithm~\ref{time elicitation} terminates in $\mathcal{A}^*$. To see the other direction, assume $R_1^*$ to be transitive. We show (1) $R_1=R_1^*$ and (2) $R_2=R_2^*$. To see (1), first choose $(a_i,a_j) \in R_1$. If $a_i=a_j$ we are done as $R_1^*$ is assumed to be reflexive. So, let $a_i \neq a_j$. If the pair $\{a_i,a_j\}$ has been presented, then $(a_i,a_j) \in R_1$ implies $(a_i,a_j) \in R_1^*$.  If $\{a_i,a_j\}$ has not been presented, there exist $a_{i_1}, \dots , a_{i_k} \in A$ such that
\begin{itemize}
    \item[i)]  $a_i=a_{i_1}~\wedge~a_j= a_{i_k}$.
    \item[ii)] For all $p=1, \dots , k-1$ the pair $\{a_{i_p},a_{i_{p+1}}\}$ has been presented.
    \item[iii)] For all $p=1, \dots , k-1$ it holds $(a_{i_p},a_{i_{p+1}}) \in R_1$.
\end{itemize}
From ii) and iii) we can conclude that it holds $(a_{i_p},a_{i_{p+1}}) \in R^*_1$ for all $p=1, \dots , k-1$. By transitivity of $R_1^*$ and i) this implies $(a_i,a_j) \in R_1^*$. To see the other direction, choose $(a_i,a_j) \in R^*_1$. If $a_i=a_j$ we are done as $R_1$ contains the diagonal of $A \times A$ by construction. So, let $a_i \neq a_j$. If $\{a_i,a_j\}$ has been presented, then $(a_i,a_j) \in R^*_1$ implies $(a_i,a_j) \in R_1$. If the pair $\{a_i,a_j\}$ has not been presented, there exist $a_{i_1}, \dots , a_{i_k} \in A$ such that properties i) and ii) hold and additionally
\begin{itemize}
    \item[iv)] $a_{i_1} R_1 a_{i_2} R_1 \dots R_1 a_{i_k}~~~\vee ~~~ a_{i_k} R_1 a_{i_{k-1}} R_1 \dots R_1 a_{i_1}$
\end{itemize}
By ii) and iv) we conclude that
\begin{itemize}
    \item[v)] $a_{i_1} R_1 a_{i_2} R_1 \dots R_1 a_{i_k}~~~\vee ~~~ a_{i_k} R^*_1 a_{i_{k-1}} R^*_1 \dots R^*_1 a_{i_1}$
\end{itemize}
By transitivity of $R_1^*$ and $(a_i,a_j) \in R^*_1$ and i) this implies
\begin{itemize}
    \item[vi)] $a_{i_1} R_1 a_{i_2} R_1 \dots R_1 a_{i_k}~~~\vee ~~~ a_{i_k} I_{R^*_1} a_{i_{k-1}} I_{R^*_1} \dots I_{R^*_1} a_{i_1}$
\end{itemize}
By construction and ii), this implies
\begin{itemize}
    \item[vii)] $a_{i_1} R_1 a_{i_2} R_1 \dots R_1 a_{i_k}~~~\vee ~~~ a_{i_k} I_{R_1} a_{i_{k-1}} I_{R_1} \dots I_{R_1} a_{i_1}$
\end{itemize}
By transitivity of $R_1$ and i) this implies $a_i R_1 a_j~\vee ~ a_j I_{R_1} a_j$ and thus $(a_i,a_j) \in R_1$.\\[.2cm]
To see (2), note that, according to the discussion of the more efficient version of time elicitation, the computed values of the consideration times coincide with the original values. Moreover, by (1), we know that there are no pairs $(a_i,a_j) \in R_1^*$ for which Algorithm~\ref{time elicitation} produces no time value $t_{ij}$. Together, this implies that Algorithm~\ref{time elicitation} produces the same relation $R_2$ as time elicitation (after having shown all pairs $\{a_i,a_j\}$), since both are defined by the same rule over the same set of time values. This allows us to conclude (2) as a direct consequence of Proposition~\ref{prop1}. 
\hfill $\square$\\
\begin{algorithm}[t]\label{time elicitation}
\DontPrintSemicolon
  
  \KwInput{consequence set $A= \{a_1, \dots , a_n\}$;}
  \KwOutput{preference system $\mathcal{A}=[A;R_1,R_2]$}
  \KwData{$R_1 = \{(a,a):a \in A\}$; $C= \emptyset$; $T=$diag$(c_{\infty})$ diagonal matrix}

 \While{$R_1 \cup R_1^{-1}\cup C\neq A\times A$}
   {
   		Sample $\{a_i,a_j\} \in A_{\{2\}}\setminus \bigl\{\{a,b\}: (a,b) \in R_1 \vee (b,a)\in R_1 \vee (a,b) \in C\bigr\}$
   		
   		\If{$a_i$ and $a_j$ are incomparable}
    {
        $R_1=R_1$\; 
        $C= C \cup \{(a_i,a_j),(a_j,a_i)\}$
    
    }
    \ElseIf{$a_i$ is strictly preferred to $a_j$}
    {
    	Set $R_1=H(R_1\cup\{(a_i,a_j)\})$\; 
        Measure $t_{ij}$ and set $T[i,j]:=t_{ij}$\; 
        \For{$(k,l) \in \{(p,q):T[p,q]=0 \wedge (a_p,a_q) \in R_1\}$}    
        { 
        	compute $t_{kl}$ using Equation~(\ref{fill up})  and set $T[k,l]:=t_{kl}$
        }
    }
    \ElseIf{$a_j$ is strictly preferred to $a_i$}
    {
    	Set $R_1=H(R_1\cup\{(a_j,a_i)\})$\; 
        Measure $t_{ji}$ and set $T[j,i]:=t_{ji}$\; 
        \For{$(k,l) \in \{(p,q):T[p,q]=0 \wedge (a_p,a_q) \in R_1\}$}    
        { 
        	compute $t_{kl}$ using Equation~(\ref{fill up}) and set $T[k,l]:=t_{kl}$
        }
    }
    \Else
    {
    	Set $R_1=H(R_1\cup\{(a_i,a_j),(a_j,a_i)\})$\;
    	Set $T[i,j]:=T[j,i]:= c_{\infty}$\;
    	\For{$(k,l) \in \{(p,q):T[p,q]=0 \wedge (a_p,a_q) \in R_1\}$}    
        { 
        		Set $T[k,l]:= c_{\infty}$
        }
    }
   }
   Define $R_2$ by setting $((a_i,a_j),(a_k,a_l)) \in R_2~~~:\Leftrightarrow~~~t_{kl}-t_{ij}\geq 0 ~\wedge~t_{ij}>0$
\caption{Efficient version of time elicitation}
\end{algorithm}
\subsection{Discussion of the assumptions underlying time elicitation} \label{discussion}
We now want to give a brief discussion of the assumptions that are required for time elicitation to work. The most important one is Assumption~\ref{time assumption} as it states how exactly consideration times and preference strength have to relate to each other. The main intuition behind this assumption is that the preference between consequences is more intense if the decision maker needs only little time for constructing a ranking of them. Consequently, it is also assumed that the maximal consideration time is attained for consequences between which the decision maker is indifferent, as indifference can be interpreted as the lowest possible preference intensity of one consequence over the other. The assumption can be given a pretty physical motivation: If the consequences to be ranked are indeed physical objects, then ranking two such objects will take more time if they are very similar to each other, since it will be harder to spot their differences. In contrast, the decision maker will be able to quickly rank very dissimilar objects (as long as these are comparable at all). For a graphic example consider a trader for jewelry pricing different gems: Given two very dissimilar gems, it will be rather easy for the expert to quickly rank them by worth even with the naked eye. In contrast, the more the two gems are alike in terms of shape, kind and look, the harder it will be for the expert to rank them by worth, since this requires time-intensive usage of special tools. 

Importantly, observe that \textit{only} Assumption~\ref{time assumption} is required for time elicitation to reproduce the decision maker's true preference system (see Proposition~\ref{time elicitation}). Thus, for judging whether this method in principle is suitable for the considered practical problem, only the adequateness of this assumption has to be verified. Assumptions~\ref{additive}~and~\ref{indi} are far more technical in nature and are not needed for the basic version of time elicitation. Instead, these two assumptions are needed for the more efficient version of time elicitation that is described in Algorithm~\ref{time elicitation} as they allow to \textit{compute} non-measured consideration times for pairs that are not directly elicited but deduced from transitivity of $R_1^*$. Clearly, compared to Assumption~\ref{time assumption}, the Assumptions~\ref{additive}~and~\ref{indi}~seem rather strong as they implicitly treat consideration times as a cardinal construct. Thus, it should be carefully thought about whether they are (at least approximately) appropriate before applying Algorithm~\ref{time elicitation} instead of the basic time elicitation procedure.

Finally, observe that the assumptions can be also discussed from the opposite side: Whenever one intends to utilize  paradata like consideration times for eliciting preference systems, certain assumptions on the relation of these and preference strength will have to be made. Assumption~\ref{time assumption} (and also~\ref{additive}~and~\ref{indi}) formalize precise conditions under which such paradata \textit{may be used} without producing wrong preference systems. Thus, implicitly, it also teaches us that a decision maker that rejects the assumptions should better not use such data for preference elicitation. For the latter case, we next present an elicitation method that avoids the usage of consideration times. 
\subsection{Procedure 2: Label elicitation} \label{pro2}
Time elicitation relies on concrete assumptions on the connection of consideration times and preference strength. Thus, this procedure should of course not be applied in situations not meeting these assumptions. Therefore, we next propose an elicitation procedure that allows for constructing an approximate version of $R_2^*$ while asking questions only about $R_1^*$. In contrast to time elicitation, this construction does not rely on data implicitly collected during the elicitation, but instead utilizes explicitly elicited \textit{labels of preference strength}. These labels are intended to provide ordinal information about preference strength. The intuition behind is very simple: To every presented pair of consequences, the decision maker assigns a label from some previously fixed set of labels. In case two presented pairs are comparable, the assigned labels will be ordered and we add the corresponding pair of pairs to the relation approximating $R_2^*$ whenever the first pair receives a strictly greater label than the latter (or both receive label $0$).

\textbf{Label elicitation} works as follows: As before, we have a finite set $A=\{a_1, \dots , a_n\}$ of consequences. We start with two empty relations $R_1 = \emptyset$ and $R_2 = \emptyset $. We then successively ask about the preferences between some (not necessarily all) pairs $(a_i , a_j) \in A \times A$, where the decision maker assigns exactly one label from the set $\mathcal{L}_r:= \{\mathbf{n}, \mathbf{c},0,1 , \dots , r\}$ to every such pair. The decision maker's labelling process can then be described by a \textit{labelling function} $\ell_r: A \times A \to \mathcal{L}_r$.

The labels from $\mathcal{L}_r$ are interpreted as follows: The higher the label from $\mathcal{L}_r \setminus \{0, \mathbf{n}, \mathbf{c}\}$ assigned to a pair $(a_i , a_j) \in A \times A$ is, the stronger is the decision maker's strict preference of $a_i$ over $a_j$. If the label $\mathbf{n}$ is assigned to $(a_i , a_j)$, this means that $a_i$ and $a_j$ are incomparable, whereas the label $0$ is interpreted as indifference between $a_i$ and $a_j$.  If the label $\mathbf{c}$ is assigned to $(a_i , a_j)$, this means that $a_i$ is strictly preferred to $a_j$, however, no statement about intensity of preference is possible.  For simplicity, we sometimes write $\ell_r^{ij}$ instead of $\ell_r((a_i,a_j))$.

The collected labels are utilized to successively build up a preference system: Whenever $\ell_r^{ij} \in \mathcal{L}_r\setminus \{\mathbf{n},0\}$, we add the pair $(a_i,a_j)$ to our relation $R_1$. If $\ell_r^{ij}=0$, we add both pairs $(a_i,a_j)$ and $(a_j,a_i)$ to our relation $R_1$, whereas if $\ell_r^{ij}=\mathbf{n}$ the relation $R_1$ remains unchanged. This procedure leaves us with a (potentially non-complete) relation $R_1$ approximating the ordinal part $R^*_1$ of the true preference system. Subsequently, we can utilize the labels of preference intensity that we collected during the procedure for also constructing an approximate version $R_2$ for the cardinal part $R_2^*$ of the decision maker's preferences. For that, we successively pick pairs of pairs $(a_i , a_j)$, $(a_k , a_l) \in R_1$ and add $((a_i , a_j),(a_k , a_l))$ to our relation $R_2$ if and only if $\ell_r^{ij}>\ell_r^{kl}$ or $\ell_r^{ij}=\ell_r^{kl} =0$.

Finally, this procedure produces a preference system $\mathcal{A}=[A,R_1,R_2]$. Importantly, note that without further assumptions this preference system does not have to coincide or even be a sub-system of the decision maker's true one. Therefore, we now give concrete assumptions under which the procedure indeed produces a meaningful preference system. 
\begin{as}\label{ass -1 proc 2}
Let $\ell_r: A \times A \to \mathcal{L}_r$ be a labelling function. It holds that
\begin{itemize}
    \item[i)]$(a_i,a_j)\in I_{R_1^*}~~ \Leftrightarrow ~~\ell_r^{ij}=0$
       \item[ii)] $(a_i,a_j)\in P_{R_1^*}~~ \Leftrightarrow ~~\ell_r^{ij} \in \mathcal{L}_r \setminus \{\mathbf{n},0\}~\wedge ~\ell_r^{ji}=\mathbf{n}$
       \item[iii)] $(a_i,a_j)\in C_{R_1^*}~~ \Leftrightarrow ~~\ell_r^{ij}= \ell_r^{ji}=\mathbf{n}$
\end{itemize}
\end{as}

Assumption~\ref{ass -1 proc 2} guarantees that, concerning the ordinal part of the preferences, the decision maker indeed uses the available labels in perfect accordance with their interpretation. In other words, it states that the labels can be used for perfectly reproducing $R_1^*$ if every pair is presented. 
\begin{as}\label{ass proc 2}
Let $\ell_r: A \times A \to \mathcal{L}_r$ be a labelling function. Then, for all $(a_i , a_j)$, $(a_k , a_l) \in R^*_1$ the following holds: 
\begin{itemize}
\item[i)]   $\ell_r^{ij}>\ell_r^{kl}~~\Rightarrow ~~~ ((a_i , a_j),(a_k , a_l)) \in P_{R_2^*}$
\item[ii)]   $\ell_r^{ij}=\ell_r^{kl} =0~~~\Rightarrow ~~~ ((a_i , a_j),(a_k , a_l)) \in I_{R_2^*}$   
\item[iii)]   $\ell_r^{ij}=\mathbf{c}~\vee ~\ell_r^{kl} =\mathbf{c}~~~\Leftrightarrow ~~~ ((a_i , a_j),(a_k , a_l)) \in C_{R_2^*}$
\end{itemize}
\end{as}

Assumption~\ref{ass proc 2} ensures that the collected labels \textit{do not conflict} with the decision maker's true cardinal part $R_2^*$. Importantly, note that in general this assumption won't be enough to reproduce the true $R_2^*$ as it does not exclude the case where $\ell_r^{ij}=\ell_r^{kl}>0$ and $ ((a_i , a_j),(a_k , a_l)) \in P_{R_2^*}$. This is intentionally: The number of labels might not be big enough to reveal the order of exchanges even if these are strictly ordered in the true preference system.
\begin{as}\label{ass 0 proc 2}
Let $\ell_r: A \times A \to \mathcal{L}_r$ be a labelling function. For all $((a_i , a_j)$, $(a_k , a_l)) \in P_{R^*_2}$ the statement $\ell_r^{ij}=\ell_r^{kl} =x \notin\{0,\mathbf{n},\mathbf{c}\}$ implies that $\{1,\dots ,r\}  \subseteq \ell_r\bigl(A \times A\bigr)$.
\end{as}

Assumption~\ref{ass 0 proc 2} guarantees that the available labels are entirely utilized whenever this is possible. In other words it forces the decision maker to label strictly ordered exchanges differently as long as there are enough labels available. Equipped with these three assumptions we can now formulate the main statement on label elicitation.
\begin{prop}\label{proplabel}
The following two statements hold true:
\begin{itemize}
    \item [i)] If, for some $r \in \mathbb{N}$, the labelling function $\ell_r: A \times A \to \mathcal{L}_r$ satisfies Assumptions~\ref{ass -1 proc 2}~and~\ref{ass proc 2}, then label elicitation produces a sub-system of the decision maker's true preference system $\mathcal{A}^*$. Particularly, it produces a consistent preference system whenever $\mathcal{A}^*$ is consistent.
    \item [ii)]  There exists $r_0 \in \mathbb{N}$ such that if labelling function $\ell_{r_0}: A \times A \to \mathcal{L}_{r_0}$  satisfies Assumptions~\ref{ass -1 proc 2},~\ref{ass proc 2}~and~\ref{ass 0 proc 2}, then label elicitation produces the decision maker's true preference system $\mathcal{A}^*$ if every pair is presented. 
\end{itemize}

\end{prop}
\textit{Proof.} i) Denote by $\mathcal{A}=[A,R_1,R_2]$ the preference system that is produced by the using labelling function $\ell_r: A \times A \to \mathcal{L}_r$. The statement $R_1 \subseteq R_1^*$ straightforwardly follows by definition of label elicitation and by Assumption~\ref{ass -1 proc 2}. To see that $R_2 \subseteq R_2^*$, choose $((a_i , a_j)$, $(a_k , a_l)) \in R_2$ arbitrarily. By definition of label elicitation, this is iff $\ell_r^{ij}>\ell_r^{kl}$ or $\ell_r^{ij}=\ell_r^{kl} =0$. By Assumption~\ref{ass proc 2}~i) and ii) this implies $((a_i , a_j),(a_k , a_l)) \in P_{R_2^*}$ or $((a_i , a_j),(a_k , a_l)) \in I_{R_2^*}$ and, thus, $((a_i , a_j),(a_k , a_l)) \in R_2^*$.\\[.15cm]
ii) Set $r_0:= |A\times A|+1=n^2+1$ and assume $\ell_{r_0}: A \times A \to  \mathcal{L}_{r_0}$  satisfies Assumptions~\ref{ass -1 proc 2},~\ref{ass proc 2}~and~\ref{ass 0 proc 2}. Let $\mathcal{A}=[A,R_1,R_2]$ be the preference system induced by $\ell_{r_0}: A \times A \to \mathcal{L}_{r_0}$ after presenting all pairs from $A \times A$. The same argument as in i) implies  $R_1 \subseteq R_1^*$ and $R_2 \subseteq R_2^*$. The statement $R^*_1 \subseteq R_1$ directly follows by Assumption~\ref{ass -1 proc 2} and the fact that every pair has been presented. Thus we have $R_1 = R_1^*$. For the cardinal part, we first show:
\begin{equation} \label{star}
    R_2^*\setminus P_{R_2}=I_{R_2^*}. 
\end{equation}
To see the direction $\subseteq$ of $(\ref{star})$, choose $((a_i , a_j),(a_k , a_l)) \in R_2^*\setminus P_{R_2}$ arbitrarily. This allows us to directly dismiss the following cases:
\begin{itemize}
     \item $\ell_{r_0}^{ij}=\mathbf{c}~\vee ~\ell_{r_0}^{kl} =\mathbf{c}$, since by Assumption ~\ref{ass proc 2} iii), this would imply $((a_i , a_j),(a_k , a_l)) \in C_{R_2^*}$ and therefore $((a_i , a_j),(a_k , a_l)) \notin R_2^*$.
      \item $\ell_{r_0}^{ij}=\mathbf{n}~\vee ~\ell_{r_0}^{kl} =\mathbf{n}$, since by Assumption~\ref{ass -1 proc 2} i) and ii) this would imply that $(a_i,a_j) \notin R_1^* ~\vee ~(a_k,a_l) \notin R_1^*$, which is impossible since $((a_i , a_j),(a_k , a_l)) \in R_2^*$.
      \end{itemize}
This leaves  $\ell_{r_0}^{ij},\ell_{r_0}^{kl} \in \{0,\dots , r_0\}$. We go on dismissing cases:
\begin{itemize}
    \item $\ell_{r_0}^{ij}>\ell_{r_0}^{kl}$, since this would imply  $((a_i , a_j),(a_k , a_l)) \in  P_{R_2}$ by construction.
    \item $\ell_{r_0}^{kl}>\ell_{r_0}^{ij}$, since by Assumption~\ref{ass proc 2} i) this would imply $((a_k , a_l),(a_i , a_j)) \in P_{R_2^*}$ contradicting our assumption that $((a_i , a_j),(a_k , a_l)) \in R_2^*$.
\end{itemize}
This leaves $\ell_{r_0}^{ij}=\ell_{r_0}^{kl} =x \notin\{\mathbf{n},\mathbf{c}\}$.\\[.1cm]
Now, note that the case $((a_i , a_j)$, $(a_k , a_l)) \in P_{R^*_2}$ is impossible: If $\ell_{r_0}^{ij}=\ell_{r_0}^{kl} =0$, we know $((a_i , a_j)$, $(a_k , a_l)) \in I_{R^*_2}$ by Assumption~\ref{ass proc 2}~ii). If $\ell_{r_0}^{ij}=\ell_{r_0}^{kl} \in \{1, \dots ,r_0\}$, by Assumption~\ref{ass 0 proc 2}, this would imply $\{1, \dots ,r_0\}  \subseteq \ell_r\bigl(A \times A \bigr) $, which is a contradiction, since a function defined on $A \times A$ can take at most $n^2$ different values.\\[.1cm]
Finally, this leaves the case $((a_i , a_j)$, $(a_k , a_l)) \in I_{R^*_2}$. \\[.15cm]
Conversely, to see the direction $\supseteq$ of $(\ref{star})$, choose $((a_i , a_j),(a_k , a_l)) \in I_{R_2^*}$ arbitrarily. Then, we have $((a_i , a_j),(a_k , a_l)) \in R_2^*$. Assume, for contradiction, that $((a_i , a_j),(a_k , a_l)) \in P_{R_2}$. By construction, this is if $\ell_{r_0}^{ij}>\ell_{r_0}^{kl}$. By Assumption~\ref{ass proc 2} i), this implies $((a_i , a_j),(a_k , a_l)) \in P_{R_2^*}$ in contradiction to $((a_i , a_j),(a_k , a_l)) \in I_{R_2^*}$. This finishes the proof of $(\ref{star})$. \\[.2cm]
We now show:
\begin{equation}\label{starstar}
    I_{R_2^*}=\Bigl\{((a_i , a_j),(a_k , a_l)): \ell_{r_0}^{ij}=\ell_{r_0}^{kl}\in \{0,\dots , r_0\}\Bigr\}
\end{equation}
To see the direction $\subseteq$ of $(\ref{starstar})$, choose $((a_i , a_j),(a_k , a_l)) \in I_{R_2^*}$ arbitrarily. As, by $(\ref{star})$, this implies $((a_i , a_j),(a_k , a_l)) \in R_2^*\setminus P_{R_2}$, we can argue as in the proof of $(\ref{star})$ that $\ell_r^{ij}=\ell_r^{kl}\in \{0,\dots , r_0\}$.\\[.1cm]
To see $\supseteq$ of $(\ref{starstar})$, choose $((a_i , a_j),(a_k , a_l)) \in \bigl\{((a_i , a_j),(a_k , a_l)): \ell_{r_0}^{ij}=\ell_{r_0}^{kl}\in \{0,\dots , r_0\}\bigr\}$ arbitrarily. Since $\ell_{r_0}^{ij}=\ell_{r_0}^{kl}\in \{0,\dots , r_0\}$ we can conclude $((a_i , a_j)$, $(a_k , a_l)) \notin P_{R^*_2}$, as seen in the proof of $(\ref{star})$. Analogous reasoning yields $((a_k , a_l)$, $(a_i , a_j)) \notin P_{R^*_2}$ and, accordingly, $((a_i , a_j)$, $(a_k , a_l)) \notin P_{R^*_2}^{-1}$. Finally, note that $\ell_{r_0}^{ij}=\ell_{r_0}^{kl} \neq \mathbf{c}$ implies $((a_i , a_j),(a_k , a_l)) \notin C_{R^*_2}$ by Assumption~\ref{ass proc 2} iii). This shows that $((a_i , a_j)$, $(a_k , a_l)) \in I_{R^*_2}$ since 
$$R_1 \times R_1 = P_{R^*_2} \cup  P^{-1}_{R^*_2} \cup I_{R_2^*} \cup C_{R_2^*}$$
where each union is disjoint. This shows $(\ref{starstar})$.\\[.15cm]
Finally, observe that $(\ref{star})$ and $(\ref{starstar})$ imply
$$R_2^*= P_{R_2} \cup \Bigl\{((a_i , a_j),(a_k , a_l)): \ell_{r_0}^{ij}=\ell_{r_0}^{kl}\in \{0,\dots , r_0\}\Bigr\}.$$ 
showing that the decision maker's true cardinal part $R_2^*$ can be fully constructed by components collected during the elicitation process.
\hfill $\square$\\[.2cm]
We now want to investigate if the efficiency of label elicitation can be further improved. Of course, one first idea would be a similar approach as followed in the context of time elicitation and to investigate in how far we can utilize transitive ordinal preferences $R_1^*$. However, due to the ordinal nature of the labels, there seems to exist no straightforward counterpart to Assumption~\ref{additive} and, therefore, no straightforward way for computing the labels of pairs that are not directly elicited but only implicitly concluded by using transitivity.\footnote{For a simple example, suppose we know $(a_i , a_j)$, $(a_j , a_k) \in R^*_1$ with labels  $\ell_r^{ij}=\ell_r^{jk}=1$. Then, by transitivity of $R_1^*$, we can conclude that $(a_i , a_k) \in R^*_1$ and, therefore, $\ell_r^{ik}\geq 1$, but nothing more than that (note again, that the labels permit no cardinal interpretation). Observe, however, that for the case that at least one of the labels $\ell_r^{ij}$ or $\ell_r^{jk}$ equals $r$ we can directly conclude that $\ell_r^{ik}$ equals $r$ as well. More generally, we can directly label all pairs with $r$ that are connected by a path of pairs containing at least one $r$ label. This, of course, should be utilized in any efficient version of label elicitation. Observe also that the improvement of proceeding as described, in general, will be higher when only few labels are considered. \label{foot7}} Clearly, one easy way out of this problem is to modify the procedure by giving label $\mathbf{c}$ to all not directly elicited pairs in $R_1$. In this way, the procedure would still produce a sub-system of the decision maker's true preference system. However, observe that this sub-system will, in general, be a strict sub-system of the one which is produced by the original version of label elicitation. Hence, such modification would result in a less informative elicitation procedure. 

For this reason, we will present another way for making label elicitation more efficient, namely a hierarchical one. The idea is very simple: After one round of elicitation is over, we restart it on equally labelled pairs. Elicitation is stopped as soon as we know that equal labelling truly originates from indifference.\footnote{Even if going for several rounds may seem less efficient at first sight, such hierarchical version has two serious advantages justifying to call it more efficient. First, as we will see in Proposition~\ref{hierprop}, it reproduces the decision maker's true preference system for any number of labels greater than one. Thus, even if the decision maker might have to answer more questions than in the basic version of label elicitation, answering these questions in accordance with the assumptions will be much less demanding if the number of labels is small. Second, a small number of labels has another serious advantage in efficiency: The fewer labels there are, the more labels can be deduced without having to present the corresponding pair (see also Footnote~\ref{foot7}).} Again, we assume that the decision maker answers in accordance with some labelling function satisfying certain assumptions. However, in addition we need to assume that the decision maker is able to adapt this labelling function to any subset of the set of all pairs $A \times A$. These adaptation process of the labelling function can be thought of as a kind of conditioning it to new information. The technical version of the assumption looks as follows:
\begin{as}\label{ass 4 proc 2}
For every $N \subseteq A \times A$ the labels on the restricted set of pairs $N$ are given with respect to a labelling function $\ell_{(N,r)}:N \to \mathcal{L}_r$ satisfying Assumptions~\ref{ass -1 proc 2},~\ref{ass proc 2} and~\ref{ass 0 proc 2} (where $\ell_r$ is replaced by $\ell_{(N,r)}$).
\end{as}
Given Assumption~\ref{ass 4 proc 2} is valid, we start elicitation with the labelling function $\ell_{(A \times A,r)}$ with some $r \geq 2$. Intuitively, the idea is picking $r$ a relatively small number so that the labelling process is not too demanding for the decision maker. After all (necessary) pairs have been shown to the decision maker, we compute the sets $A_x:=\ell_{(A \times A,r)}^{-1}(\{x\})$ for all $x \in\{1 , \dots , r\}$. If either $r \geq |A \times A|$ or (at least) one of the sets $A_x$  is empty, we can stop elicitation. Otherwise, we restart elicitation on $A_x$ with the labelling function $\ell_{(A_x,r)}$ for every $x \in\{1 , \dots , r\}$. We then compute the sets $A_{(x,y)}:=\ell_{(A_{(x)},r)}^{-1}(\{y\})$ for all $x,y \in\{1 , \dots , r\}$. If, for some $x^*$ fixed, either $r \geq |A_{x^*}|$ or (at least) one of the sets $A_{(x^*,y)}$, where $y \in\{1 , \dots , r\}$,  is empty, we can stop elicitation for those pairs with first label $x^*$. Again, we restart elicitation on the sets $A_{(x,y)}$ with  $\ell_{(A_{(x,y)},r)}$ and so on.\\[.15cm]
\indent In general, the \textbf{hierarchical version} of label elicitation works as follows: First, we set $A_{()}:=A \times A$. Then, for $p \in \mathbb{N}$ and $\underline{x}_p=(x_1, \dots ,x_p) \in \{1, \dots ,r\}^p$ chosen such that the elicitation has not terminated in $A_{(x_1)}, \dots, A_{(x_1, \dots , x_{p-1})}$, we recursively define 
\begin{equation}
    A_{\underline{x}_p}:=\ell_{(A_{\underline{x}_{p-1}},r)}^{^{-1}} (\{x_p\}).
\end{equation}
The elicitation procedure can be stopped for pairs in $A_{(x_1, \dots , x_{p-1})}$ after round $p$ if either $r \geq |A_{(x_1, \dots , x_{p-1})}|$ or (at least) one of the sets $A_{\underline{x}_p}$, where $x_p \in \{1 , \dots ,r\}$,  is empty.\\[.15cm]
\indent After the  procedure ended, every pair $(a_{i},a_j)$ can be associated with a \textit{label history}, namely the vector of labels $h^{ij}_t \in \mathcal{L}_r^t$ that have been given to it in the $t$ elicitation rounds it was involved. Importantly, observe that if one of the labels $\mathbf{n}$, $\mathbf{c}$ or $0$ is contained in the history $h^{ij}_t$ of $(a_{i},a_j)$, then this directly implies $t=1$.  This is true for two reasons: (1) Elicitation goes in round two only on the sets $A_x$ for $x \in \{1, \dots, r\}$. (2) The $k$th component of $h^{ij}_t$ must be contained in $\{1, \dots ,r\}$ whenever its $(k-1)$th component is contained in $\{1, \dots ,r\}$. Here, fact (1) follows by construction and fact (2) is an immediate consequence of Assumption~\ref{ass 4 proc 2}.   

\begin{figure}[ht]
    \centering
    \includegraphics[scale=.11]{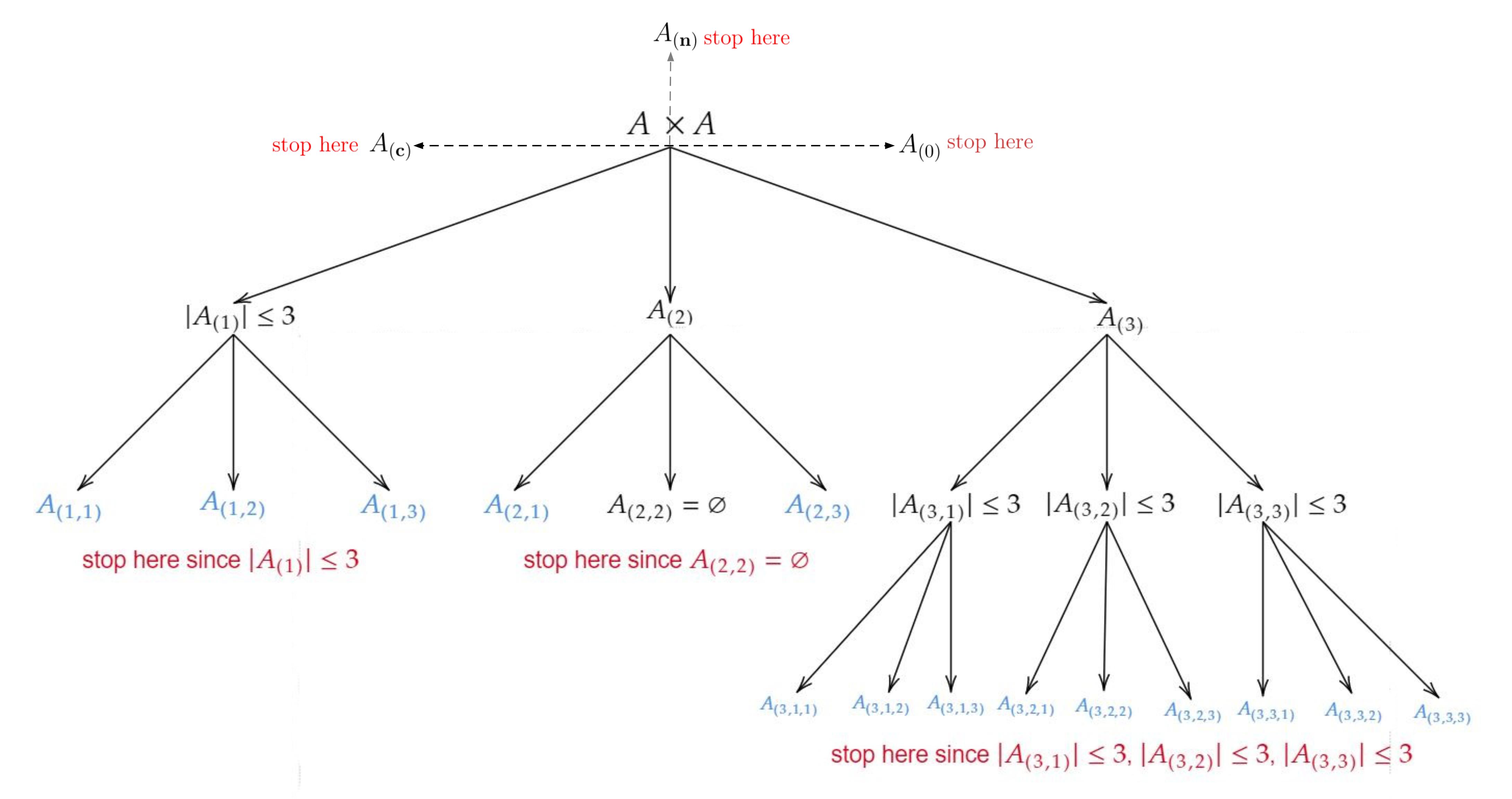}
    \caption{A schematic example for the hierarchical version of label elicitation with $r=3$. For the labels $\mathbf{n}$, $\mathbf{c}$ and $0$ the elicitation is terminated after round one. For the labels in $\{1 ,\dots ,r\}$ we keep on eliciting until either there are enough labels to (potentially) label all pairs differently or the available labels are not fully needed. In both cases it is possible to distinguish strict preference and indifference between pairs. After the procedure terminated, relation $R_1$ is constructed by collecting all pairs with first label in $\mathcal{L}_3 \setminus \{\mathbf{n}\}$, whereas $R_2$ is constructed by adding all combinations of pairs in the same blue set to $I_{R_2}$ and all combinations of pairs to $P_{R_2}$ for which the first pair ends in a blue set that lies right to the one of the second pair.}
    \label{tree}
\end{figure}

Further, observe that in general the label histories of distinct pairs might be of different dimension. The label histories are used to construct the relations $R_1$ and $R_2$ by the following rules:
\begin{itemize}
    \item $(a_i,a_j)\in R_1~~:\Leftrightarrow ~~h_t^{ij}[1]\geq 0~~\vee ~~h_t^{ij}[1]= \mathbf{c}$ (where $x[j]$ is the $j$th component of $x$) and
    \item $((a_i , a_j),(a_k , a_l))\in R_2~~:\Leftrightarrow ~~h^{ij}_{t_1}$ is lexicographically greater or equal than $ h^{kl}_{t_2}$.\footnote{Here, we slightly abuse notation. Precisely, we add $((a_i , a_j),(a_k , a_l))$ to $R_2$ iff $h^{ij}_{t_1}=h^{kl}_{t_2} \notin\{\mathbf{n},\mathbf{c}\}$ or if $h^{ij}_{t_1}$ is strictly greater on the first component they differ. Denote this relation by $\geq_L$. Observe that, by construction, this relation indeed orders all pairs with label histories not containing labels from $\{\mathbf{n},\mathbf{c}\}$.\label{lexico} }
\end{itemize}
A compact presentation of the hierarchical version of label elicitation is given in Algorithm~\ref{hierlabel}. A graphical illustration of the procedure for a small example is given in Figure~\ref{tree}. The following proposition states that Algorithm~\ref{hierlabel} indeed produces the decision maker's true preference system given Assumption~\ref{ass 4 proc 2} holds true.
\begin{prop} \label{hierprop}
Let Assumption~\ref{ass 4 proc 2} hold true. For $n=|A|$ consequences and $r \geq 2$ labels, Algorithm~\ref{hierlabel} terminates in $\mathcal{A}^*$ after at most $\max\{1,\lceil\tfrac{n^2-r}{r-1}\rceil+1\}$ elicitation rounds.
\end{prop}
\textit{Proof.} We first show that \textit{if} Algorithm~\ref{hierlabel} terminates, then it terminates in $\mathcal{A}^*$. To see this, assume the algorithm terminated in a preference system $\mathcal{A}=[A,R_1,R_2]$. The equation $R_1=R_1^*$ follows by the exact same argument as in the proof of Proposition~\ref{proplabel} ii), since the construction of $R_1$ depends only on the first elicitation round. Thus, it remains to show $R_2=R_2^*$. Let $((a_i,a_j),(a_k,a_l)) \in R_2$. This, by construction,  implies $h_{t_1}^{ij}\geq_L h^{kl}_{t_2}$ and, therefore, none of the labels in the histories of $(a_i,a_j)$ and $(a_k,a_l)$ is $\mathbf{n}$ or $\mathbf{c}$. Hence, we know $((a_i,a_j),(a_k,a_l)) \in R^*_2$ or $((a_k,a_l),(a_i,a_j)) \in R^*_2$ by Assumptions~\ref{ass -1 proc 2} iii) and~\ref{ass proc 2} iii). We distinguish three cases:
\\[.1cm]
\textit{Case 1:} $h_{t_1}^{ij}= h^{kl}_{t_2}=0$ (and thus $t_1=t_2=1$). This immediately implies $((a_i,a_j),(a_k,a_l)) \in I_{R^*_2}$ by Assumption~\ref{ass proc 2} ii) for the function $\ell_{(A \times A ,r)}$.
\\[.1cm]
\textit{Case 2:} $h_{t_1}^{ij}= h^{kl}_{t_2}$ (and thus $t_1=t_2$). Set $x:=h_{t_1-1}^{ij}$ \\[.1cm]
\textit{Sub-case 2.1:} $|A_x|\leq r$. Assume, for contradiction, that $((a_k,a_l),(a_i,a_j)) \in P_{R^*_2}$. Then, since $\ell_{(A_x,r)}$ satisfies Assumption~\ref{ass 0 proc 2} according to Assumption~\ref{ass 4 proc 2}, we know that $h_{t_1}^{ij}[t_1]= h^{kl}_{t_2}[t_1]$ implies that $\{1, \dots,r\} \subseteq\ell_{(A_x,r)}(A_x)$. This is a contradiction since $A_x$ contains at most $r$ elements and $(a_k,a_l)$ and $(a_i,a_j)$ are equally labelled. Since we know that $(a_k,a_l)$ and $(a_i,a_j)$ are comparable w.r.t. $R_2^*$ in at least one direction, this implies $((a_i,a_j),(a_k,a_l)) \in R^*_2$.  \\[.1cm]
\textit{Sub-case 2.2:} $A_{(x,y)}= \emptyset$ for at least one $y \in \{1, \dots , r\}$. Again assume that $((a_k,a_l),(a_i,a_j)) \in P_{R^*_2}$. This implies that $y \notin\ell_{(A_x,r)}(A_x)$ yielding a contradiction to Assumption~\ref{ass 0 proc 2} for $\ell_{(A_x ,r)}$ since $h_{t_1}^{ij}[t_1]= h^{kl}_{t_2}[t_1]$. By the same argument as above we conclude that  $((a_i,a_j),(a_k,a_l)) \in R^*_2$.
\\[.15cm]
\textit{Case 3:} There exists $t^* \in \{1, \dots, \min\{t_1,t_2 \}\}$ with  $h_{t_1}^{ij}[t]= h^{kl}_{t_2}[t]$ for all $t <t^*$ and  $h_{t_1}^{ij}[t^*]> h^{kl}_{t_2}[t^*]$. Let $x$ be the vector containing the first $t^*-1$ components of the label history of $h_{t_1}^{ij}$ and $h^{kl}_{t_2}$. Since the function $\ell_{(A_x ,r)}$ satisfies Assumption~\ref{ass proc 2} i), we can directly conclude that $((a_i , a_j),(a_k , a_l)) \in P_{R_2^*}$, since
$\ell_{(A_x ,r)}((a_i , a_j))=h_{t_1}^{ij}[t^*]> h^{kl}_{t_2}[t^*]=\ell_{(A_x ,r)}((a_k , a_l))$. \\[.15cm]
Thus, we have $((a_i,a_j),(a_k,a_l)) \in R^*_2$ in every possible case. \\[.15cm]
Conversely, assume that $((a_i,a_j),(a_k,a_l)) \in R^*_2$. By Assumptions~\ref{ass -1 proc 2} iii) and~\ref{ass proc 2} iii) this implies that $h_{t_1}^{ij}$ and $h^{kl}_{t_2}$ do not contain the label $\mathbf{n}$ or $\mathbf{c}$ and, therefore, that either $h_{t_2}^{kl} \geq_L h^{ij}_{t_1}$ or $h_{t_1}^{ij} \geq_L h^{kl}_{t_2}$ holds.  Assume, for contradiction, that  $(h_{t_2}^{kl}, h^{ij}_{t_1}) \in P_{\geq_L}$. Then there  exists $t^* \in \{1, \dots, \min\{t_1,t_2 \}\}$ with  $h_{t_1}^{ij}[t]= h^{kl}_{t_2}[t]$ for all $t <t^*$ and  $h_{t_1}^{ij}[t^*]< h^{kl}_{t_2}[t^*]$.  Let $x$ be the vector containing the first $t^*-1$ components of the label history of $h_{t_1}^{ij}$ and $h^{kl}_{t_2}$. Since the function $\ell_{(A_x ,r)}$ satisfies Assumption~\ref{ass proc 2} i), we can directly conclude that $((a_k , a_l),(a_i , a_j)) \in P_{R_2^*}$. Contradiction. This yields  $h_{t_1}^{ij} \geq_L h^{kl}_{t_2}$ and, therefore, $((a_i,a_j),(a_k,a_l)) \in R_2$.\\[.15cm]
Remains to show that the Algorithm terminates after at most $\max\{1,\lceil\tfrac{n^2-r}{r-1}\rceil+1\}$ elicitation rounds. This is easy to see: First, note that for $r > n^2$ the algorithms terminates after elicitation round one, since one of the sets $A_x$, with $x \in \{1, \dots ,r\}$, needs to be empty. So, let $2 \leq r \leq n^2$. Next, observe that under the worst case scenario the relation $R_1^*$ is a linear order (otherwise we `lose' the incomparable and indifferent pairs already after round one). Now, note that the worst case occurs, if after each round one of the sets on which elicitation needs to be restarted contains the maximal possible number of elements. This happens if identical label is given to the maximal possible number of pairs.  As none of the labels might remain unused in any round (otherwise elicitation is terminated), the maximal number of  pairs with identical label is $d-(r-1)$, where $d$ denotes the number of remaining pairs in round $k$. As $A \times A$ contains $n^2$ pairs, we can compute the number of rounds after which this maximal set contains at most $r$ elements by finding the smallest integer $k$ such that $n^2-k(r-1)\leq r$. This gives $k=\lceil\tfrac{n^2-r}{r-1}\rceil$. By construction of the procedure we then need one more round yielding the upper bound $\lceil\tfrac{n^2-r}{r-1}\rceil+1$.
\hfill $\square$\\[.2cm]
\begin{algorithm}[h]\label{hierlabel}
\DontPrintSemicolon
  
  \KwInput{consequence set $A= \{a_1, \dots , a_n\}$; number of labels $r$}
  \KwOutput{preference system $\mathcal{A}=[A;R_1,R_2]$}
Set $\mathcal{B}=\{A \times A\}$\;
Set $h^{ij}=()$ for all $i,j\in \{1, \dots ,n\}$\;
 \While{$|\mathcal{B}|>0$}{
   \For{$N \in \mathcal{B}$}
   {
   \For{$(a_i,a_j)\in N$}
   {Set $h^{ij}=(h^{ij},\ell^{ij}_{(N,r)})$\;}
   \If{$|N| \leq r$}{Set $\mathcal{B}=\mathcal{B}\setminus\{N\}$}
    \If{$N_x = \emptyset$ for some $x \in \{1 , \dots ,r\}$}{Set $\mathcal{B}=\mathcal{B}\setminus\{N\}$}
   
   }
  Set $\mathcal{B}=\{N_x: N \in \mathcal{B}, x \in \{1 , \dots , r\}\}$ 
}
Construct $R_1$ by setting $(a_i,a_j)\in R_1~~~:\Leftrightarrow ~~~h^{ij}[1]\geq 0 ~~\vee ~~h^{ij}[1]= \mathbf{c}$\;
Construct $R_2$ by setting $((a_i,a_j),(a_k,a_l))\in R_2~~~:\Leftrightarrow ~~~h^{ij}\geq_L h^{kl}$

\caption{Hierarchical version of label elicitation}
\end{algorithm}
\subsection{Improving the procedures via statistically guided pair selection schemes} \label{data}

In this section we briefly describe how both procedures can be further improved. Instead of randomly sampling a pair of consequences in every elicitation step one can choose the next pair in a more principled way. For this we assume that we have additional information in form of a sample of the elicited preference systems of previous users. Moreover, we assume that the decision maker who is currently elicited has a preference system that is similar to that of the previous users, or at least similar to a subset of the preference systems of the previous users. The idea for improving the process of elicitation is now that we can statistically guide with this assumption which pair of consequences should be elicited in the next step. This is achieved by using a statistical method that predicts which next pair one should choose to make the elicitation procedure information efficient, i.e., to reduce the number of queried pairs. There are in principle two ways to proceed: i) One can make the choice of the next pair of consequences by incorporating the exact used decision rule to guide the eliciting process in such a way that a decision can already be made at a very early step of the procedure. ii) One can base the choice of the next pair of consequences only on the preference system and without reference to the decision rule, but with the same aim of reaching a decision very early in the process of elicitation. This would have the advantage that one does not need to know the decision rule of the decision maker. In the sequel we will only discuss the second approach in more detail. The first approach will be briefly discussed in the outlook.
One very simple heuristic for choosing the next pair of alternatives in every step is to choose that pair $(a_i,a_j)$ for which the proportion of previous users who ranked $a_i$ before $a_j$ is closest to one. Another more advanced approach would be to use any kind of prediction procedure for every pair $(a_i,a_j)$ that uses all already elicited pairs $(a_k,a_l)$ as features and all not elicited pairs $(a_i,a_j)$ as outcomes for which one has to make a prediction. Then, similar to the first approach one can choose that pair $(a_i,a_j)$ for which the prediction is most certain. These two approaches will be briefly illustrated within a short simulation given in Example \ref{bsp2}, where we explicitly discuss a multimodal model of partial rankings. As a prediction method for new pairs to elicite we use there the method of subgroup discovery.
%
\section{Applying the procedures to decision making in complex information settings} \label{decmak}
Our focus so far was on efficiently eliciting the decision maker's true preference system $\mathcal{A}^*$.~Apart from the one about non-elicited preferences, no uncertainty was involved (see also Section~\ref{prel}). We now investigate what can be learned by our elicitation procedures in decision making under uncertainty, i.e.~when the decision maker has to choose among \textit{acts} $X_i:S \to A$ taking values in the set $A$. Any attempt for decision making then clearly should be based on the available information on \textit{both} the relations $R_1^*$ and $R_2^*$ \textit{and} the mechanism generating the states in $S$.
\subsection{Modelling decision making in complex information settings}\label{bm}
We assume the decision maker's preferences on $A$ are adequately described by the preference system $\mathcal{A}^*=[A,R^*_1,R^*_2]$, where still $A=\{a_1, \dots ,a_n\}$ is a finite set of consequences. However, the consequence that a specific decision produces now depends on which state of nature from $S=\{s_1 , \dots ,s_m\}$ occurs. The decision maker thus is faced with a finite set of acts $\mathcal{G} =\{X_1 , \dots , X_k\} \subseteq A^S$ out of which it may be chosen. For a schematic visualization of the problem see Table~\ref{bm-table}.
\begin{table}[h]
\begin{center}
\begin{tabular}{c|ccc}
\hline\noalign{\smallskip}
 & $\mathbf{s_1}$ & $\cdots$ & $\mathbf{s_m}$\\
\noalign{\smallskip}
\hline
\noalign{\smallskip}
$\mathbf{X_1}$ & $X_1(s_1) \in A$ &$\cdots$ & $X_1(s_m) \in A$  \\
$\vdots$ & $\vdots$ & $\cdots$ & $\vdots$  \\
$\mathbf{X_k}$ & $X_k(s_1) \in A$ & $\cdots$ & $X_k(s_m) \in A$  \\
\hline
\end{tabular}
\end{center}
\caption{The basic model of finite decision theory. The acts $X_i$ take values in the set $A$ and, therefore, should be ordered by utilizing the information encoded in the decision maker's preference system $\mathcal{A}^*$. As the concrete consequence the decision maker receives after choosing $X_i$ depends on the state of nature $s$, any meaningful ordering of the acts should also incorporate the information about the mechanism generating the states of nature.}
\label{bm-table}
\end{table}

Moreover, we assume there is also information on the mechanism generating the states $s \in S$. This information is assumed to be characterized by a polyhedral \textit{credal set} $\mathcal{M}$ of probability measures on the set $S$.\footnote{Credal sets are widely accepted models for situations under complex uncertainty. For the general theory of credal sets, interval probabilties, or most generally imprecise probabilities, see, e.g.,~\citet{levi,Walley.1991,Weichselberger.2001}. For  recent introductions to the theory  see~\citet{Augustin} and \citet{Bradley:2019}.} Any element $\pi \in \mathcal{M}$ is assumed to be an equally plausible candidate for the true probability and there thus is no meaningful way for further reducing the uncertainty about the states. Note that this assumption allows for very general uncertainty models ranging from perfect probabilistic information (in this case $\mathcal{M}$ reduces to a singleton) to complete ignorance (in this case $\mathcal{M}$ is the set of all probabilities on $S$, the so called \textit{vacuous} set).\footnote{Importantly, note that the focus of the present paper is on eliciting preference systems rather than credal sets (or more generally imprecise probabilistic models). Precisely, we assume the uncertainty model for the states of nature to be externally given in what follows. For works focusing on the elicitation of imprecise probabilities see, e.g., \citet{protocol, erik, miranda}. For a survey see~\citet{Smithson.2014}.} In particular, credal sets allow to model decision making problems under severe uncertainty about the states.

The full problem is then given by the \textit{decision system} $(\mathcal{A}^*, \mathcal{G}, \mathcal{M})$ and optimal acts are usually determined by a \textit{choice function} ch$:2^{\mathcal{G}} \to 2^\mathcal{G}$ satisfying ch$(\mathcal{W}) \subseteq \mathcal{W}$ for all $\mathcal{W} \in 2^{\mathcal{G}}$. This choice function should best possibly utilize the information about the preferences encoded in $\mathcal{A}^*$ and the information about the the states encoded in $\mathcal{M}$.\footnote{For a recent introduction to the theory of choice functions see \citet[Chapter 2]{chambers2016revealed}. For surveys on choice functions in the context of imprecise probabilities see, e.g.,~\cite{troffaes_ijar,Huntley.2014,bradley}. For computational aspects see, e.g.,~\citet{compitip,Jansen2017,festschrift}.} A number of choice functions for decision systems is discussed in~\cite{jsa2018}. The ones most relevant for the present paper are briefly recalled in the following Section~\ref{rule}.

We can now precisely formulate the main question of this section: Can we find satisfying solutions to some decision system without fully specifying the decision maker's preference system? In other words: Suppose after some steps of an elicitation procedure we have produced the preference system $\mathcal{A}$ being a strict subset of $\mathcal{A}^*$. When do we have the same set of optimal acts ch$(\mathcal{G})$ no matter if we base the decision on $(\mathcal{A}, \mathcal{G}, \mathcal{M})$ or $(\mathcal{A}^*, \mathcal{G}, \mathcal{M})$? Clearly, the answer to this depends on the choice function and can not be answered this general. Thus, in Section~\ref{dmel} we investigate it for two specific choice functions for decision systems. Before that, we briefly recall these.
\subsection{Two different notions of dominance} \label{rule}
\cite{jsa2018} introduce various ways for defining choice functions for decision systems. We want to restrict our analysis to only two of these, namely such choice functions relying on \textit{generalized expectation intervals} and such relying on \textit{global expectation dominance}. Before we can come to these choice functions, we need one more definition.
\begin{mydef}\label{gei}
Let $(\mathcal{A}, \mathcal{G}, \mathcal{M})$ be a decision system based on some $\delta$-consistent $\mathcal{A}=[A,R_1,R_2]$, where $\delta \in [0,1)$. For $X \in \mathcal{G}$, we define the generalized expectation interval for granularity $\delta$ by
\begin{equation}
\mathbb{E}_{\mathcal{D}_{\delta}}(X):=\Bigr[\underline{ \mathbb{E}}_{\mathcal{D}_{\delta}}( X) ,  \overline{ \mathbb{E}}_{\mathcal{D}_{\delta}}( X) \Bigl]:=\Bigr[\inf\limits_{(u, \pi) \in \mathcal{D}_{\delta}}  \mathbb{E}_{\pi}( u\circ X) , \sup\limits_{(u, \pi) \in \mathcal{D}_{\delta}}  \mathbb{E}_{\pi}(  u\circ X) \Bigl] \label{e_lower}
\end{equation}
where we have set $\mathcal{D}_{\delta} :=\mathcal{N}^{\delta}_{\mathcal{A}} \times \mathcal{M}$. For simplicity, the value $\mathbb{E}_{\mathcal{D}}(X):=\mathbb{E}_{\mathcal{D}_{0}}(X)$ is called the generalized expectation interval of $X$.
\end{mydef}
Based on the generalized expectation interval just defined, we can now define one of the two choice functions mentioned before. The other considered choice function does rely on the idea of point-wise expectation dominance rather than generalized expectation intervals. The following definition is based on~\citet[Definitions 5 and 7]{jsa2018}. 
\begin{mydef} \label{cf}
 Let $(\mathcal{A}, \mathcal{G}, \mathcal{M})$ be some decision system that is based on some $\delta$-consistent (for i)) or consistent (for ii)) preference system $\mathcal{A}=[A,R_1,R_2]$. Then:
 \begin{itemize}
     \item[i)] For $\delta \in [0,1)$, $\delta$-interval dominance is defined by the choice function $D_{\mathcal{A}}:2^{\mathcal{G}} \to 2^{\mathcal{G}} $ with
\begin{equation}
   D_{\mathcal{A}}(\mathcal{W}):=\Bigr\{Y \in \mathcal{W}:\forall X\in \mathcal{W} \text{ it holds }\underline{ \mathbb{E}}_{\mathcal{D}_{\delta}}( Y) \geq   \overline{ \mathbb{E}}_{\mathcal{D}_{\delta}}( X)\Bigl\}.
\end{equation}
\item[ii)] $\mathcal{A}|\mathcal{M}$-dominance is defined by the choice function P$_{\mathcal{A}}:2^{\mathcal{G}} \to 2^{\mathcal{G}} $ with
\begin{equation}
   P_{\mathcal{A}}(\mathcal{W}):=\Bigr\{Y \in \mathcal{W}:\forall X\in \mathcal{W},u \in \mathcal{U}_{\mathcal{A}}, \pi \in \mathcal{M} \text{ it holds }  \mathbb{E}_{\pi}(u \circ Y) \geq \mathbb{E}_{\pi}(u \circ X) \Bigl\}.
\end{equation}
 \end{itemize}

\end{mydef}
Some words on interpretation: The choice function $D_{\mathcal{A}}$ can be thought of as a generalization of the classical interval dominance rule known from the theory of imprecise probabilities to the case where also the utility function is not precisely specified. Importantly, observe that the generalized interval expectation and thus also the choice function $D_{\mathcal{A}}$ are only well-defined if $A$ possesses minimal and maximal elements with respect to $R_1$. This is due to the fact that the definition of the set $\mathcal{N}^{\delta}_{\mathcal{A}}$ makes explicit use of this assumption (see also Section~\ref{prel}). This gives rise to the following subtlety in the elicitation context: Even if the decision maker's true ordinal relation $R_1^*$ satisfies this assumption, it might not be known prior to the elicitation procedure. Consequently, the choice function  $D_{\mathcal{A}}$ can only be applied if we know in advance one of the decision maker's worst and best consequences. Pragmatically, this shortfall can be solved by artificially adding such elements to the consequence set prior to the elicitation procedure (see also Footnote~\ref{foot6}). 

In contrast, the choice function $P_{\mathcal{A}}$ completely avoids assumptions on the existence of minimal and maximal elements of $A$ with respect to $R_1$ and, therefore, is applicable to an even more general class of decision problems. It generalizes first order stochastic dominance to the cases of imprecise probabilistic models and partial cardinal preference information. Further generalization of first order stochastic dominance to imprecise probabilities can be found in \citet{denoeux},~\citet{couso} or \citet{montemir}.
\subsection{Eliciting optimal decisions} \label{dmel}
How can we utilize the elicitation procedures from Section~\ref{elisec} for defining more efficient solution strategies for decision making problems \textit{under uncertainty}? The idea is very simple: Once the elicitation procedure is started, we evaluate the selected choice function after each (or some fixed number of) elicitation step(s) and terminate the procedure as soon we can find an optimal act for the first time. That is, we terminate elicitation as soon as we find an optimal act in the decision system that is based on the preference system elicited so far.

More specifically, we arrive at the following: Let $(\mathcal{A}^*, \mathcal{G}, \mathcal{M})$ denote a decision system based on some consistent decision maker's true preference system $\mathcal{A}^*=[A,R^*_1,R^*_2]$. Moreover, denote by $\mathcal{A}_1, \mathcal{A}_2, \dots ,\mathcal{A}_k, \dots$ the preference system that is produced after step $1,2,\dots, k ,\dots$ of either time or label elicitation or one of its variants (which one doesn't matter for the following considerations). Suppose we want to compute an optimal act from $\mathcal{G}$ with respect to $C_{\mathcal{A}^*} \in \{D_{\mathcal{A}^*},P_{\mathcal{A}^*}\}$. Then after each elicitation step $k$ we can check whether it holds that
$$X \in C_{\mathcal{A}_k}(\mathcal{G})$$
for every $X \in \mathcal{G}$ separately and stop the procedure for the smallest $k^*$ for which $ C_{\mathcal{A}_{k^*}}(\mathcal{G}) \neq \emptyset$. This check can be done by, e.g., using linear programming theory.\footnote{For linear programming-based algorithms for checking the condition $X \in C_{\mathcal{A}_k}(\mathcal{G})$ for fixed $k$, see~\citet[Prop. 5]{jsa2018}~for $C_{\mathcal{A}_k}=D_{\mathcal{A}_k}$ and~\citet[Prop. 3]{jsa2018}~for $C_{\mathcal{A}_k}=P_{\mathcal{A}_k}$.} For any act $X \in C_{\mathcal{A}_{k^*}}(\mathcal{G})$ we then conclude that it is optimal with respect to the selected choice function also in the true decision system $(\mathcal{A}^*, \mathcal{G}, \mathcal{M})$. Thus, we stop elicitation as soon as the selected choice function produces a non-empty choice set for the first time and conclude any act in this choice set is also optimal for the original (and, in general, not yet fully elicited) decision system. \textit{Given such conclusion is valid}, this creates the possibility that the considered decision maker can solve the faced problem under uncertainty by only answering (a potentially moderate number of) simple ranking questions about $R_1^*$. Observe the connection to demand (III)$^{'}$ mentioned in the introduction of the paper: Even if the considered decision maker's true preference structure is perfectly cardinal, proceeding as just described allows us to specify only those parts of it that are important for the concrete decision to be made. The following proposition states that this conclusion \textit{is indeed valid}.
\begin{prop} \label{prop_decmak}
Let $C_{\mathcal{A}} \in \{D_{\mathcal{A}},P_{\mathcal{A}}\}$ and $\delta \in [0,1)$. Further, let $(\mathcal{A}^*, \mathcal{G}, \mathcal{M})$ be based on some consistent (if $C_{\mathcal{A}}=P_{\mathcal{A}}$) or $\delta$-consistent (if $C_{\mathcal{A}}=D_{\mathcal{A}}$) preference system $\mathcal{A}^*$. Suppose $\mathcal{A}^*$ is elicited by either time or label elicitation producing $\mathcal{A}_1, \mathcal{A}_2, \dots ,\mathcal{A}_k, \dots$ after step $1,2,\dots, k ,\dots$ of elicitation. Assume the decision maker satisfies Assumption~\ref{time assumption} for time elicitation or Assumptions~\ref{ass -1 proc 2}~and~\ref{ass proc 2} for label elicitation, respectively. Then for all $k$ it holds that $X \in C_{\mathcal{A}_k}(\mathcal{G})$ implies $X \in C_{\mathcal{A}^*}(\mathcal{G})$.
\end{prop}
\textit{Proof.} Assume $X \in C_{\mathcal{A}_k}(\mathcal{G})$ for some $k$. By assumptions of the proposition and Corollary~\ref{cor1} (in case that time elicitation is applied) or Proposition~\ref{proplabel} i) (in case that label elicitation is applied), respectively, we can conclude that $\mathcal{A}_k$ is a sub-system of $\mathcal{A}^*$. Since by assumption $\mathcal{A}^*$ is consistent, we know that $\mathcal{U}_{\mathcal{A}^*}$ is non-empty. Hence, by  Corollary~\ref{cor1} or Proposition~\ref{proplabel} i), respectively, we can conclude that $\mathcal{A}_k$ is consistent and, therefore, that $\mathcal{U}_{\mathcal{A}_k}$ is non-empty as well. Moreover, by applying the definition of a sub-system one easily verifies that  $\mathcal{U}_{\mathcal{A}_k} \supseteq \mathcal{U}_{\mathcal{A}^*}$ (this is due to the fact that a function $u:A \to [0,1]$ weakly representing $\mathcal{A}^*$ has to satisfy all the constraints induced by $\mathcal{A}_k$ and, in case of inequality, even more additional constraints). We now distinguish two cases:\\[.1cm]
\textit{Case 1:} $C_{\mathcal{A}}=P_{\mathcal{A}}$. Then $X \in P_{\mathcal{A}_k}(\mathcal{G})$ and it thus holds that  $\mathbb{E}_{\pi}(u \circ X)\geq\mathbb{E}_{\pi}(u \circ Y)$ for every $Y \in \mathcal{G}$, $u \in \mathcal{U}_{\mathcal{A}_k}$ and $\pi \in \mathcal{M}$. Since we have  $\mathcal{U}_{\mathcal{A}_k} \supseteq \mathcal{U}_{\mathcal{A}^*}$, we may exchange  $\mathcal{U}_{\mathcal{A}_k} $ by $\mathcal{U}_{\mathcal{A}^*}$ and the statement holds still true. This implies $X \in P_{\mathcal{A}^*}(\mathcal{G})$.\\[.1cm]
\textit{Case 2:} $C_{\mathcal{A}}=D_{\mathcal{A}}$. Then $X \in D_{\mathcal{A}_k}(\mathcal{G})$. By $\delta$-consistency and $\mathcal{U}_{\mathcal{A}_k} \supseteq \mathcal{U}_{\mathcal{A}^*}$ we know $\mathcal{N}^{\delta}_{\mathcal{A}_k}\supseteq \mathcal{N}^{\delta}_{\mathcal{A}^*} \neq \emptyset$. This implies the following inequalities (where $Y \in \mathcal{G}$ is chosen arbitrarily):
\begin{equation*}
\inf\limits_{\mathcal{N}^{\delta}_{\mathcal{A}^*} \times \mathcal{M}}  \mathbb{E}_{\pi}( u\circ X) \geq \inf\limits_{\mathcal{N}^{\delta}_{\mathcal{A}_k} \times \mathcal{M}}  \mathbb{E}_{\pi}( u\circ X)\geq \sup\limits_{\mathcal{N}^{\delta}_{\mathcal{A}_k} \times \mathcal{M}}  \mathbb{E}_{\pi}( u\circ Y)\geq \sup\limits_{\mathcal{N}^{\delta}_{\mathcal{A}^*} \times \mathcal{M}}  \mathbb{E}_{\pi}( u\circ Y).
\end{equation*}
This yields $X \in D_{\mathcal{A}^*}(\mathcal{G})$, completing the proof.
\hfill $\square$\\[.2cm]
Proposition~\ref{prop_decmak} guarantees that any act $X$ that is in the choice set $C_{\mathcal{A}_k}(\mathcal{G})$ of the choice function $C_{\mathcal{A}}$ based on the preference system $\mathcal{A}_k$ elicited so far, will remain in the choice set $ C_{\mathcal{A}^*}(\mathcal{G})$ of the same choice function based on the decision maker's true preference system $\mathcal{A}^*$. However, if we are interested in the whole choice set $ C_{\mathcal{A}^*}(\mathcal{G})$ the statement loses its bite: It might be the case that $X \notin C_{\mathcal{A}_k}(\mathcal{G})$ but $X \in C_{\mathcal{A}^*}(\mathcal{G})$. Hence, in this case it is not possible to terminate the corresponding elicitation procedure in advance.
\section{Two stylyzed application examples} \label{aex}
\begin{figure}[h]
    \centering
    \includegraphics[scale=.7]{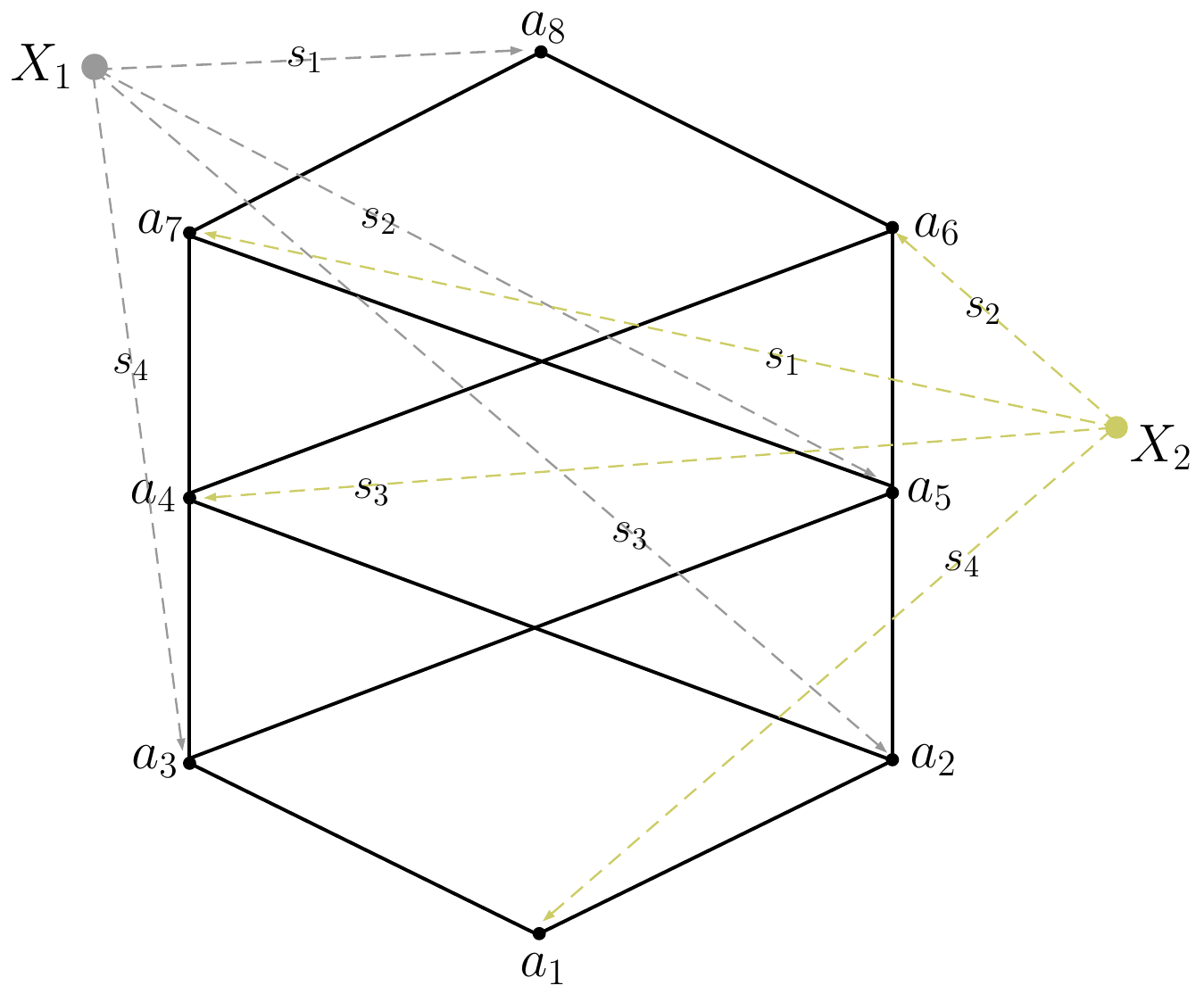}
    \caption{Hasse diagram of the ordinal part $R_1^*$ for Example~\ref{Example1}. The dotted lines illustrate the consequences the available acts $X_1$ and $X_2$ from Table~\ref{table_ex1} attain under the different states of nature.}
    \label{ex1}
\end{figure}

In this section we will demonstrate and discuss the proposed methods based on two examples. These examples are chosen to be of increasing generality. We start by a very simple example in which the uncertainty about the states is described by a single probability measure. Further, we demonstrate that the example can easily be modified to a situation under severe uncertainty modelled by a comparative credal set on the states of nature.
\begin{bsp}\label{Example1}\end{bsp}
Let $A=\{a_1, \dots , a_8\}$ be a set containing eight different consequences. Moreover, suppose the decision maker's true consistent preference system $\mathcal{A}^*$ on $A$ is defined as follows: The (anti-symmetric) relation $R_1^*$ is induced by the Hasse diagram depicted in Figure~\ref{ex1}.\\

\noindent The relation $R_2^*$ is given as the transitive hull of 
$$e_{31} P_{R_2^*} e_{52}  P_{R_2^*}  e_{74}  P_{R_2^*} e_{21} I_{R_2^*} e_{64} I_{R_2^*} e_{42} I_{R_2^*} e_{86}  P_{R_2^*}  e_{87} P_{R_2^*} e_{53} P_{R_2^*}e_{75} P_{R_2^*} e_{65} P_{R_2^*} e_{43} $$
where $e_{ij}:=(a_i,a_j)$ for all $i,j \in \{1, \dots , 8\}$. Assume that the decision maker is faced with the simple decision problem with only two available acts given in Table~\ref{table_ex1}. 
\begin{table}[ht]
\begin{center}
\begin{tabular}{c|cccc}
\hline\noalign{\smallskip}
 & $\mathbf{s_1}$ & $\mathbf{s_2}$ & $\mathbf{s_3}$ & $\mathbf{s_4}$\\
\noalign{\smallskip}
\hline
\noalign{\smallskip}
$\mathbf{X_1}$ & $a_8$ &$a_5$  & $a_2$ & $a_3$  \\
$\mathbf{X_2}$ & $a_7$ &$a_6$  & $a_4$ & $a_1$  \\
\hline
\end{tabular}
\end{center}
\label{table_ex1}
\caption{A compact representation of the decision problem in Example~\ref{table_ex1}. }
\end{table}
Further, let the uncertainty about the states be characterized by a classical probability $\pi$, i.e. we have $\mathcal{M}=\{\pi\}$, where $\pi$ is the uniform distribution on $S$. Assume elicitation is done by using label elicitation in its basic version. The decision maker labels by a function $\ell_5:A \times A \to \mathcal{L}_5$ that satisfies the Assumptions~\ref{ass -1 proc 2},~\ref{ass proc 2} and~\ref{ass 0 proc 2}. Moreover, assume the first four elicitation steps look as follows:\footnote{Observe that the labels given to these four pairs are in perfect accordance with the Assumptions~\ref{ass -1 proc 2},~\ref{ass proc 2} and~\ref{ass 0 proc 2}. Note further that these pairs were not randomly sampled, but chosen with respect to the simple heuristics of \textit{presenting pairs of consequences different acts yield under the same state}. Clearly, there is no obvious way of generalizing such heuristics to the case of more than two acts available. Still the example demonstrates that it is possible to solve complex decision problems by answering very few simple ranking questions.}
\begin{center}
\begin{tabular}{|c|c|c|}
\hline\noalign{\smallskip}
 Elicitation step & Presented pair & Label of the pair \\
\noalign{\smallskip}
\hline
\noalign{\smallskip}
1 & $(a_8,a_7)$ &$\ell_5^{87}=2$  \\
2 & $(a_6,a_5)$ &$\ell_5^{65}=1$  \\
3 & $(a_3,a_1)$ &$\ell_5^{31}=3$  \\
4 & $(a_4,a_2)$ &$\ell_5^{42}=2$  \\
\hline
\end{tabular}
\end{center}
Thus, after four elicitation steps we arrive the preference system $\mathcal{A}_4=[A, R_1,R_2]$, where
\begin{itemize}
    \item $P_{R_1}=\{(a_8,a_7),(a_6,a_5),(a_3,a_1),(a_4,a_2)\}$ and
    \item $P_{R_2}=\{(e_{31},e_{42}),(e_{31},e_{65}),(e_{31},e_{87}),(e_{87},e_{65}),(e_{42},e_{65})\}$.
\end{itemize}
Then, for every $u \in \mathcal{U}_{\mathcal{A}_4}$, we can go on computing (where $u_i := u(a_i)$):
\begin{equation*}
4\cdot(\mathbb{E}_{\pi}(u \circ X_1)-\mathbb{E}_{\pi}(u \circ X_2)) = \underbrace{(u_8-u_7)-(u_6-u_5)}_{>0,~since~(e_{87},e_{65})\in P_{R_2}}+\underbrace{(u_3-u_1)-(u_4-u_2)}_{>0,~since~(e_{31},e_{42})\in P_{R_2}} >0
\end{equation*}
and thus $\mathbb{E}_{\pi}(u \circ X_1)>\mathbb{E}_{\pi}(u \circ X_2)$ for every $u \in \mathcal{U}_{\mathcal{A}_4}$. This gives $X_1 \in P_{\mathcal{A}_4}(\mathcal{G})$. By applying Proposition~\ref{prop_decmak} for $C_{\mathcal{A}}= P_{\mathcal{A}}$ we can conclude that $X_1 \in P_{\mathcal{A}^*}(\mathcal{G})$. Thus, $X_1$ is $\mathcal{A}|\mathcal{M}$-dominant. Note that this was concluded by asking only four simple ranking questions.

The example just discussed can easily be modified in two different directions: First, `time elicitation' could be used instead of `label elicitation'. For example, if one were to assume that for the pairs $(a_8,a_7),(a_6,a_5),(a_3,a_1)$ and $(a_4,a_2)$ the consideration times $t_{87}=0.3s$, $t_{65}=0.5s$, $t_{31}=0.2s$ and $t_{42}=0.35s$ were measured, the rest of the example would remain unchanged: Still a decision could be made after asking four simple ranking questions.

Second, instead of a precise probability, the credal set 
$$\mathcal{M}_c=\Bigl\{\pi: \pi(\{s_1\}) \geq \pi(\{s_2\}) \geq\pi(\{s_4\}) \geq \pi(\{s_3\})\Bigr\}$$  
could be used as an uncertainty model. One then easily verifies that for all $u \in \mathcal{U}_{\mathcal{A}_4}$ and $\pi \in \mathcal{M}_c$ the following inequality holds (where $\pi_j:= \pi(\{s_j\})$):
\begin{equation*}
\mathbb{E}_{\pi}(u \circ X_1)-\mathbb{E}_{\pi}(u \circ X_2) = \underbrace{\pi_1 (u_8-u_7)-\pi_2(u_6-u_5)}_{>0,~since~(e_{87},e_{65})\in P_{R_2}~\wedge ~\pi_1 \geq \pi_2}+\underbrace{\pi_4 (u_3-u_1)-\pi_3(u_4-u_2)}_{>0,~since~(e_{31},e_{42})\in P_{R_2}~\wedge ~\pi_4 \geq \pi_3} >0
\end{equation*}
The exact same argument as in the case of a precise $\pi$ then yields that $X_1$ is $\mathcal{A}|\mathcal{M}_c$-dominant which again was concluded by asking only four ranking questions.
\begin{bsp}\label{bsp2}\end{bsp}
We now again slightly modify the situation discussed in Example~\ref{Example1} by adding one more act $X_3$ to the set of available acts. The rest of the example remains unchanged. In particular, we again use the choice function $P_{\mathcal{A}}$ as our criterion of optimality. A compact description of the extended problem is given in Table~\ref{table_ex2}. 

\begin{table}[ht]
\begin{center}
\begin{tabular}{c|cccc}
\hline\noalign{\smallskip}
 & $\mathbf{s_1}$ & $\mathbf{s_2}$ & $\mathbf{s_3}$ & $\mathbf{s_4}$\\
\noalign{\smallskip}
\hline
\noalign{\smallskip}
$\mathbf{X_1}$ & $a_8$ &$a_5$  & $a_2$ & $a_3$  \\
$\mathbf{X_2}$ & $a_7$ &$a_6$  & $a_4$ & $a_1$  \\
$\mathbf{X_3}$ & $a_1$ &$a_4$  & $a_6$ & $a_7$  \\
\hline
\end{tabular}
\end{center}

\caption{A compact representation of the decision problem in Example~\ref{bsp2}. }\label{table_ex2}
\end{table}

With this example we want to illustrate how additional data from previous elicitations of previous decision makers can be used to statistically guide the elicitation process in order to make the elicitation more efficient for a new decision maker. For the sake of comparison, we firstly evaluate how many steps of elicitation are needed for making a decision between $X_1,X_2$ and $X_3$ when the pairs of consequences are presented randomly. Observe that the above heuristic can not be directly adopted for more than two acts. (Note that in this example acts $X_2$ and $X_3$ are equivalent w.r.t. first order stochastic dominance already for an empty $R_1^*$, thus the heuristic of Example \ref{Example1} would also need only four steps, but this is only due to the example.) 

In a simulation of $100$ random elicitations, it turned out that the median number of pairs to present for making a decision was $14$ (the interquartile range ($IQR$) was $4$). Now, for a statistically guided elicitation we first need some data from previous elicitation processes.  For this we simulated partial orders on a ground set of eight outcomes according to a simple statistical model. Concretely, we use the Mallows model \citep{Mallow1957} adapted to the generation of arbitrary partial orders (i.e., not only total orders) with the partial order $R_1^*$ from Example~\ref{Example1} as a mode and simulated partial orderings according to their distance to this mode:

$$P(R_1=r_1) = c\cdot \exp\left\{-\frac{1}{\lambda} \cdot d\left(r_1,R_1^*\right)\right\}.$$

Here, $c$ is a normalization constant, $\lambda >0$ is a spread parameter and $d$ is the distance between two partial orders, measured by the cardinality of their symmetric difference (i.e., the Hamming distance/Manhattan distance). Additionally, we assume that the currently elicited decision maker's preference system is in accordance with Example \ref{Example1}. 
Because already for eight alternatives the number of partial orders (i.e., reflexive, transitive and antisymmetric relations) is very high (concretely, there are $431,723,379$ partial orders), we decided not to enumerate all partial orders. Instead we directly sampled a number ($n=100$ and $n=250$, respectively) of orders. Therefore, we modify the algorithm in  \cite{ganter2011random} for generating closures from an arbitrary closure operator.\footnote{For  the  notion  of  a  closure  operator  and  the  language  of  Formal  Concept  Analysis,  which  is  used  in  \cite{ganter2011random}, see \cite{ganter2005formal}.} We use the family of all partial orders on a ground set with a fixed number of elements. In a first step we decomposed this space: For a fixed linear order $L$  we looked at the closure system of all antisymmetric binary relations that can be linearly extended by $L$. The corresponding closure operator maps each such binary antisymmetric relation on its  transitive hull. After randomly sampling an order $L$, the algorithm randomly samples a set of binary antisymmetric relations $R$ that can be extended by $L$. By computing the transitive hull and the transitive reduction it obtains the probability that the  transitive hull of $R$ is sampled. This algorithm can also be applied in larger decision problems with far more than eight outcomes where an explicit enumeration of all partial orders is impossible.

For choosing the next pair to present to the decision maker, in a first approach, we used a simple heuristic: For every pair we computed the proportion of orders in the simulated data that contained the given pair. Then we presented the pairs according to this proportion, in decreasing order. With this, the median of the number of needed pairs was reduced to $12$ ($IQR = 1$) for a training data set of $n=100$ partial orders. (Here $\lambda$ was set to $1$.) This reduction seems to be not too big, but note that we have only three incomparable pairs in $R_1^*$ and thus nearly every arbitrarily sampled pair gives some information and therefore reduces the space $\mathcal{U}_{\mathcal{A}_k}$. Note further that for our simple heuristic, at every step we did not use that pairs of the current decision maker that were already elicited. The reason for this is that within our simple Mallows-type model, beyond transitivity, already elicited pairs essentially do not contain any information w.r.t.~non-elicited pairs.\footnote{The reason for this is an exponential form together with a simple addition of non-coinciding pairs within the metric, which essentially leads to an independence model.}

Using the already elicited pairs for predicting the next pair does only make sense in a model where the already elicited pairs do contain some information about the non-elicited pairs. Such models are more difficult to establish, especially if one is interested in models that are acceptable approximations of empirically reasonable elicitation situations. Here, we will only sketch one very simple bimodal model where one can see more or less analytically that using the already elicited pairs will make the  elicitation process more efficient compared to the more simple heuristic from above, which does not use the already elicited pairs.

Like above, in a first step we use the Mallows model, but with the modification that with probability $1/2$ we sample not the original order, but instead we sample the corresponding inverse order (where $R_1$ is replaced by $R_1^{-1}$). Additionally, for ease of analysis, we use here a total order as a mode and we only sample total orders. With this, because of symmetry, the simple heuristic that uses only the proportions of pairs in the data set seemingly does not help in choosing a good next pair for the elicitation.\footnote{This is true at least for the first step of the elicitation process. Note that for the further steps, in contrast to a random elicitation where the process of choosing the next pair is exchangeable, for the case of an elicitation that uses the proportions of pairs, this is not the case. This can still by accident lead to a better or worse performance of this elicitation, compared to a random elicitation.}

On the other hand (at least if the spread parameter $\lambda$ is not too big) if one did already elicit some pairs, then one will know with some certainty if the decision maker one elicits has the original order or the inverse order. In a simulation with $100$ replications and  with a training data set of $n=250$ orders (randomly generated independently for every replication) we got a median number of $11$ elicitation steps. ($IQR = 1$, $\lambda$ was set here to $0.5$.) For the prediction of the new pair we used the methodology of subgroup discovery (see, e.g., \cite{herrera2011overview,atzmueller}):

For every possible target pair $(a,b)$ that was not already elicited, we computed that subgroup of orderings that contain all pairs of the already elicited decision maker and that at the same time has the purest distribution w.r.t.~pair $(a,b)$. Concretely, we maximized the Piatetsky-Shapiro quality measure $n(p-p_0)$ where $n$ is the size of the subgroup, $p$ is the proportion of pairs $(a,b)$ in the subgroup and $p_0$ is the proportion of pairs $(a,b)$ in the whole group.

The obtained results show that it could be useful to consider the already elicited pairs for statistically guiding the elicitation process. (For a random elicitation the median was $15$ steps ($IQR = 6$). For elicitation according to the observed proportions of pairs we got a median of $15$ with $IQR=5.25$.) Of course, the used model is still very simple, but in principle it can be adopted to distributions of partial orders with more than two modes. Then, also for more than two modes (and of course, also for partial orders), with enough data, if enough pairs are already elicited, one can in fact informatively predict, to which mode the decision maker presumably belongs. This is actually a realistic and flexible model under which the statistically guided elicitation has in fact some merits.

\section{Summary and outlook} \label{outlook}
In this paper we presented two different approaches for efficiently eliciting some decision maker's preference system. \textit{Time elicitation} utilizes consideration times for constructing the cardinal part of the preference system, whereas \textit{label elicitation} directly collects labels of preference strength for doing so. Both methods are efficient in the sense that learning the decision maker's preference system requires asking only a small amount of simple ranking questions about its ordinal part. For both methods, we gave conditions under which these questions produce (approximations of) the true preference system of the decision maker under consideration. Moreover, we discussed different approaches for further improving the two procedures: For time elicitation, we presented an algorithm that, given the decision maker has transitive ordinal preferences and two further assumptions are satisfied, already produces the decision maker's true preference system if only ranking questions on a subset of the set of all pairs are asked. For label elicitation, we presented a hierarchical modification of the original procedure that guarantees reproducing the true preference system for any number of labels greater than one while being cognitively less demanding for the considered decision maker. 

For both procedures (and their modifications), we proposed a method for a \textit{statistically guided} selection of the next pair to present in every elicitation step relying on prior knowledge from previous elicitation procedures. The main idea behind this method then showed its power in the context of decision making under severe uncertainty: Since criteria for finding optimal acts depend on the set of utility functions that are compatible with the true preference system, in every step the procedure selects a pair that is expected to most effectively shrink this set. More generally speaking, we demonstrated (in particular in the examples from the previous section) how the different variants of the proposed procedures can be used for solving complexly structured decision making problems without having to fully specify the decision maker's preference system. Thus, we indeed managed to develop a framework for decision making that adequately accounts for all three demands (I), (II)$^{'}$ and (III)$^{'}$ announced in the introduction.

There are several promising directions for future research. We now briefly sketch those among them that seem most important in our eyes:
\begin{itemize}
    \item \textit{Improved prediction of promising pairs:} The \textit{statistically guided} procedure for presenting effective new pairs from Section~\ref{data} utilizes only the ordinal parts of previously elicited preference systems. Also incorporating their cardinal parts could help speeding up elicitation by making prediction even more effective. But there is also space for improvement in a non-data context.
    
    In label elicitation it turned out to be hard utilizing transitive ordinal preferences as there seems to be no obvious rule for labelling pairs that follow from transitivity (see also Footnote~\ref{foot7}). Given a concrete decision system and choice function, one way out of this problem could be the following: First elicit only on pairs that are not implied by transitivity. Afterwards, elicit the exact labels of those implied pairs that produce ordinal constraints that are sharply valid for optimal solutions of the optimization problem induced by the decision system and the choice function. Proceeding like this can be viewed as an \textit{optimization-driven} prediction procedure for selection among the implied pairs. 
    
    In the context of time elicitation we argued that the additivity assumption (\ref{additive}) is very strong. However, after the elicitation is done and the decision can be made under this strong assumption, one can -- in an additional validation process -- check all constraints of the involved linear program that are sharply valid for the obtained solution of the linear program. If some of the constraints are not supported by the preferences of the decision maker, one can further elicit the preferences of the decision maker until all relevant sharp constraints are explicitly supported by the decision maker. With such a post-elicitation procedure one can in fact ensure the validity of all involved constraints and thus one does not really rely on the strong additivity-assumption (\ref{additive}). Instead, one only uses this assumption to guide the elicitation.
    
    \item \textit{Mixing hierarchical and non-hierarchical procedures:} In Section~\ref{pro2} we presented two methods for label elicitation: one hierarchical, the other single-staged. For making elicitation more efficient, these could be combined in different ways: First, it seems promising to find some criterion that in each elicitation step allows for deciding whether it is more effective to elicit a new (yet not elicited) pair or to re-elicit some pair on a finer scale. Second, for every pair, the decision maker could be allowed to flexibly choose also the scale (i.e. the number of labels) on which this concrete pair is evaluated. In this way the decision maker could locally express preference intensity on different scales for every pair.
    
    \item \textit{Statistically investigating the behaviour of the prediction of new pairs:} For the prediction of pairs to present in the elicitation procedure, we used the methodology of subgroup discovery. In every step we constrained the training set to that partial orderings that are in accordance with all already elicited pairs of the decision maker. This means that in every step the training data set is getting smaller and at the same time the covariates for the prediction of new pairs are getting less complex. A thorough analysis of how this effects a possible over- or underfitting of the prediction procedure would be of high interest here. Fortunately, for the case of subgroup discovery, for example an explicit and computationally feasible analysis in terms of Vapnik-Chervonenkis theory (see, e.g., \cite{vapnik2006estimation}) is possible. Because of the close connection between formal concept analysis and subgroup discovery (cf., e.g., \cite{Boley2009,boley2010formal}), this does also allow for the application of regularization strategies within this discrete setting, where, for instance, the ideas developed in \cite[p. 26 ff]{epub40416} could proof fruitful (cf., also \cite{Boley2017,Boley2018}).
    
     Another interesting point would be a more explicit analysis of the prediction method under still more complex but analytically manageable models.
 \item \textit{Explicitly incorporating the choice function into the prediction of new pairs:} We did not yet explicitly discuss the explicit incorporation of the choice function into the prediction method. One way of incorporation would be to make in every step of the elicitation a prediction of the whole preference system of the decision maker. Then one could proceed as if this prediction was the true underlying preference system and choose that chain of further elicitation steps which leads to the most effective elicitation. This would of course be computationally very hard, but at least for example greedy heuristics would  presumably be easy to establish. 
    \item \textit{Investigate other choice functions:} In Section~\ref{decmak} we demonstrated that under $\delta$-interval dominance or $\mathcal{A}|\mathcal{M}$-dominance optimal decisions sometimes can be determined without fully specifying the decision maker's preference system. The main reason for this turned out to be that the sets of utility functions that are produced after the different elicitation steps are nested. It seems very relevant to investigate and develop stopping rules for choice functions that do not have this property.
\end{itemize}
\section*{Acknowledgements}
We thank both anonymous reviewers and the editor for valuable comments that helped to improve the paper. In particular, we are thankful for suggesting to add a `summary for a non-mathematical audience'. We thank the participants of the ISIPTA '21 and IFORS 2021 conferences for valuable comments on and discussions about some early-stage ideas connected to the concepts presented in this paper. Further, we thank Marc Johler for providing parts of the code used in Example~\ref{bsp2} of Section~\ref{aex}.~Hannah Blocher and Georg Schollmeyer gratefully acknowledge the financial and general support of the LMU Mentoring program. 
%
\bibliographystyle{apalike}
\bibliography{references.bib}
\end{document}